%% file: main.tex
\documentclass[UTF8]{article} 
\usepackage{iclr2023_conference,times}

\usepackage{microtype}
\usepackage{amsmath}
\usepackage{amssymb}
\usepackage{booktabs}
\usepackage{colortbl}
\usepackage{CJKutf8}
\usepackage[utf8]{inputenc}
\definecolor{lightgray}{rgb}{0.9,0.9,0.9}
\usepackage{caption}
\usepackage{multicol}
\usepackage{wrapfig}
\usepackage{subcaption}
\usepackage{xcolor}
\usepackage{graphicx}
\usepackage{setspace}
\usepackage{hyperref}
\usepackage{url}
\usepackage{multirow}
\usepackage{colortbl}
\usepackage{tabularx}
\usepackage{blindtext}
\usepackage{pgfplots}
\pgfplotsset{compat=1.18} 
\usepackage{tikz}
\usetikzlibrary{er,positioning,bayesnet}
\usepackage[inline]{enumitem}
\usepackage{makecell}
\usepackage{tipa}
\usepackage{siunitx}
\usepackage{tocloft}
\usepackage{listings}
\usepackage[raster,skins]{tcolorbox} 
\usepackage{xltabular}
\usepackage{hyperref}
\usepackage[framemethod=tikz]{mdframed}
\usepackage{graphicx}
\surroundwithmdframed[
  hidealllines=true,
  innerleftmargin=0pt,
  innertopmargin=0pt,
  innerbottommargin=0pt]{lstlisting}
\lstnewenvironment{response}[1][] 
 {\lstset{
 columns=fullflexible,
  breakautoindent=false, breakindent=0pt, breaklines, linewidth=8cm, #1}}
 {}

\newcommand{\method}{\textbf{ALIVE}\xspace}

\newlength\savewidth\newcommand\shline{\noalign{\global\savewidth\arrayrulewidth
  \global\arrayrulewidth 1pt}\hline\noalign{\global\arrayrulewidth\savewidth}}

\definecolor{url_color}{RGB}{113, 187, 179}

\hypersetup{
    colorlinks=true,
    linkcolor=black,
    urlcolor=url_color,
}

\title{ALIVE: Animate Your World with Lifelike Audio-Video Generation}
\author{
\textbf{Bytedance ALIVE Team}
}

\iclrfinalcopy
\begin{document}

\maketitle

\begin{abstract}
Video generation is rapidly evolving towards unified audio-video generation.  In this paper, we present \method, a generation model that adapts a pretrained Text-to-Video (T2V) model to Sora-style audio-video generation and animation.  In particular, the model unlocks the Text-to-Video\&Audio (T2VA) and  Reference-to-Video\&Audio (animation) capabilities compared to the T2V foundation models.
To support the audio-visual synchronization and reference animation, we augment the popular MMDiT architecture with a joint audio-video branch which includes TA-CrossAttn for temporally-aligned cross-modal fusion and UniTemp-RoPE for precise audio-visual alignment. Meanwhile, a comprehensive data pipeline consisting of audio-video captioning, quality control, etc,  is carefully designed to collect high-quality finetuning data. Additionally, we introduce a new benchmark to perform a comprehensive  model test and comparison.  After continue pretraining and finetuning on million-level high-quality data, 
\method demonstrates outstanding performance, consistently outperforming open-source models and matching or
surpassing state-of-the-art commercial solutions. With detailed recipes and benchmarks, we hope \method  helps the community develop audio-video generation models more efficiently.  Official page: \url{https://github.com/FoundationVision/Alive}.

\end{abstract}

\vspace{13pt}

\begin{figure}[htbp]
    \centering
    \includegraphics[width=0.95\textwidth, trim=0 80 0 94, clip]{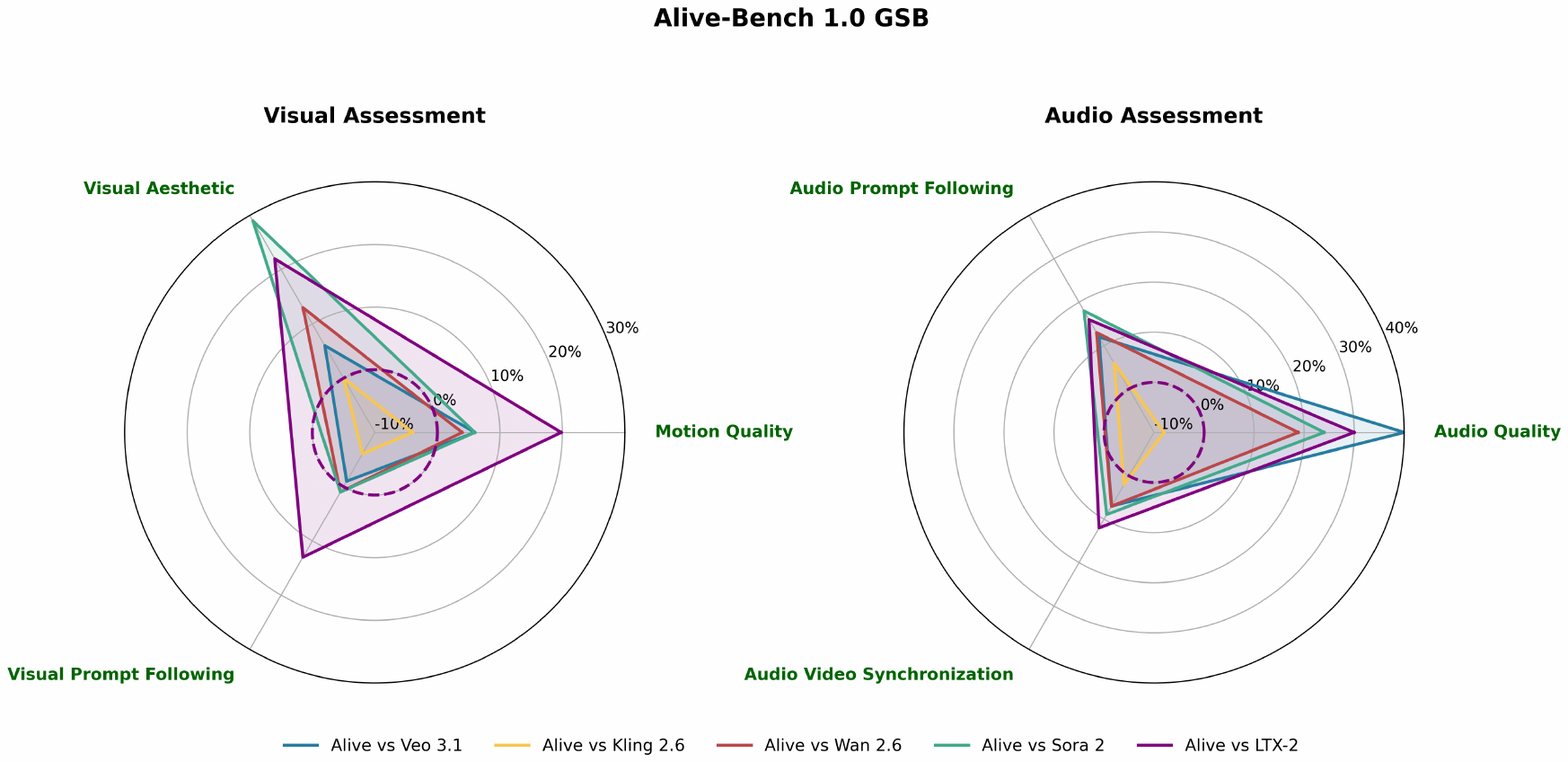}
    \caption{
        \textbf{Left: }Human evaluation win rates (GSB) of \method compared to Veo 3.1, Kling 2.6, Wan 2.6, Sora 2 and LTX-2 on Alive-Bench 1.0 across six dimensions: Motion Quality, Visual Aesthetic, Visual Prompt Following, Audio Quuality, Audio Prompt Following and Audio Video Synchronization. Alive-Bench 1.0 covers a wide range of scenarios, including single-person speech, multi-people conversations, sports, daily activities, animals, means of transportation, surreal scenes, etc.
    }
    \label{fig:common_eval}
\end{figure}

\clearpage






\newpage

\begin{figure}[htbp]
    \centering
    \includegraphics[width=\textwidth]{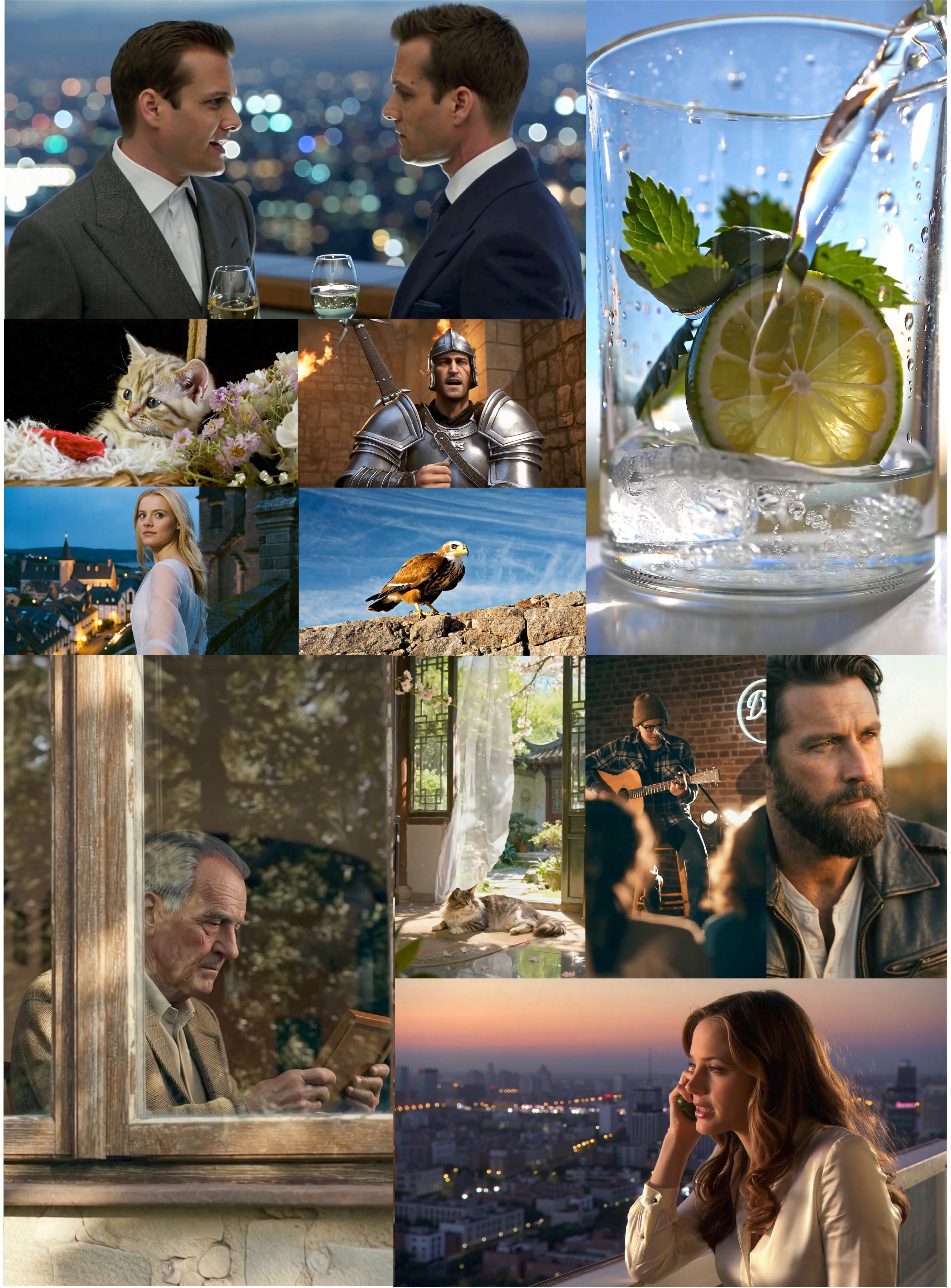}
    \caption{
        T2V samples generated by \method. \method is capable of generating 1080p videos at arbitrary aspect ratios, delivering high levels of aesthetic quality, realism, and motion fidelity, while simultaneously supporting both T2VA, I2VA and R2VA tasks.
    }
    \label{fig:demo}
\end{figure}

\clearpage


\tableofcontents
\clearpage

\input{content/1_intro}

\input{content/2_model_arch}

\input{content/3_data}

\input{content/4_recipe}

\input{content/5_role_playing_animate}

\input{content/6_bench}

\input{content/8_conclusion}

\input{content/9_authors}

\bibliography{main}
\bibliographystyle{iclr2023_conference}
\clearpage

\end{document}

%% file: content/1_intro.tex
\section{Introduction}
\label{sec:intro}

The field of video generation has witnessed remarkable progress in recent years, with significant academic and industry advances propelled by models such as Wan~\citep{wan2025wan}, StepVideo~\citep{ma2025step}, HunyuanVideo~\citep{kong2024hunyuanvideo}, and CogVideoX~\citep{cogvideox}, Veo~\citep{veo32025}, Kling~\citep{kuaishou2024kling}. 
At present, there is a clear trend of video generation evolving towards audio-video generation for immersive media creation.  New models such as Veo 3.1, Wan 2.6, and Sora 2 showcase notable progress in unified audio-video synthesis, enabling natural speech lip-sync and dynamic background sounds. The integration of audio and video generation enables more vivid and expressive creative content, further powering a wide range of applications.

However, extending a high-quality video generation model to support synchronized audio generation still presents several significant challenges:
(1) In multilingual audio generation, it remains challenging to produce speaker timbre that sounds natural and avoids a mechanical or artificial quality. (2) Additionally, achieving accurate audio-visual synchronization is difficult, with high error rates especially evident in multi-speaker lip-sync scenarios. (3) Existing public technical reports do not clearly describe how to scale previous visual-only data curation pipelines to include high-quality audio-video data cleaning and captioning. (4) Methods for evaluating the performance of audio-visual models are still lagging behind, with no widely accepted standards or comprehensive benchmarks.

To address these challenges, we propose \method to advance the popular text-to-video model to support  audio-video generation and animation with two main modules: Joint Audio-Video DiT and Cascade Audio-Video Refiner.  Multi-stage training strategies are adopted to generate  1080p  high-resolution audio-augmented videos  from 5 to 10 seconds with optional reference image input. In this work, we  instantiates our model based on the pretrained  Waver 1.0~\citep{zhang2025waver} to perform joint audio-video generation. In summary, we introduce the following key improvements and contributions:

\begin{itemize} [leftmargin=*,labelsep=0.6em]

\item \textbf{Joint Audio-Video Modeling. } We propose \method, a joint generation architecture that seamlessly integrates Audio and Video DiTs via an extended ``Dual Stream + Single Stream" paradigm. To resolve temporal granularity mismatches, we introduce \textbf{UniTemp-RoPE} and \textbf{TA-CrossAttn}, which map heterogeneous latents into a shared continuous temporal coordinate system, enforcing physical-time alignment for synchronized audio-visual generation.

\item \textbf{Comprehensive Audio-Video Data Pipelines. }
Going beyond conventional visual quality filtering, our work introduces a comprehensive data pipeline for joint audio-visual generation. It performs dual-quality filtering on both audio and video modalities, and employs a joint `visual + audio' keyword labeling system to associate a single visual object with its diverse range of audio events, enabling a more sophisticated level of audio-visual data balancing. Furthermore, we optimize and correct the Subject-Speech correspondence in multi-person scenarios, significantly enhancing character identity consistency and accuracy.

\item \textbf{Alive-bench 1.0. }
We introduce a comprehensive benchmark, Alive-Bench 1.0 for joint audio-visual generation that evaluates model performance along three complementary axes, which are motion quality, visual aesthetic, visual prompt following, audio quality, audio prompt following and audio video synchronization, covering 22 detail metrics.

\item \textbf{Role-Playing Animate. } 
We introduce a cross-pair pipeline and a unified-editing-based reference augmentation pipeline to decouple identity from static appearance, effectively mitigating copy-paste bias. Furthermore, we develop a multi-reference conditioning mechanism with a dedicated temporal offset, enabling the model to treat reference images as persistent identity anchors rather than temporal frames, thus achieving superior identity consistency and motion dynamics.

\end{itemize}

\method demonstrates superior performance on the Alive-Bench 1.0, achieving state-of-the-art performance across various domains including visual asethetic, audio prompt following and audio-video synchronization. By natively supporting reference animiation within our audio-video synthesis framework, we empower everyone to animate their world with lifelike audio-visual content and foster further advancements in multi-modal generative research.

%% file: content/2_model_arch.tex
\section{Model Architecture}
\label{sec:model-arch}

Building upon the foundation of Waver 1.0~\citep{zhang2025waver}, \method advances the paradigm from silent video synthesis to joint audio-visual generation. We retain the robust Wan2.1-VAE~\citep{wan2025wan} for efficient video latent compression and the powerful dual-encoder system (Flan-T5-XXL~\citep{t5} and Qwen2.5-32B-Instruct~\citep{qwen2.5}) to ensure superior text understanding and prompt adherence.
The core architecture of \method is redesigned as a Joint Audio-Video DiT, based on rectified flow Transformers~\citep{esser2024sd3}. This unified framework seamlessly integrates a well-designed Audio DiT with the Video DiT~\citep{zhang2025waver}. To facilitate deep modality interaction, we extend the ``Dual Stream + Single Stream" architecture, where the audio and video latents interact primarily through TA-CrossAttn (Temporally-Aligned Cross-Attention) mechanisms. To address the challenge of precise temporal synchronization between visual latents and audio latents, we propose a novel UniTemp-RoPE (Unified Temporal Rotary Positional Embedding) in TA-CrossAttn. This design enforces strict temporal alignment across modalities, enabling the generation of high-fidelity, synchronized audio-visual content. 

\begin{figure}[htbp]
    \centering
    \includegraphics[width=1.0\textwidth]{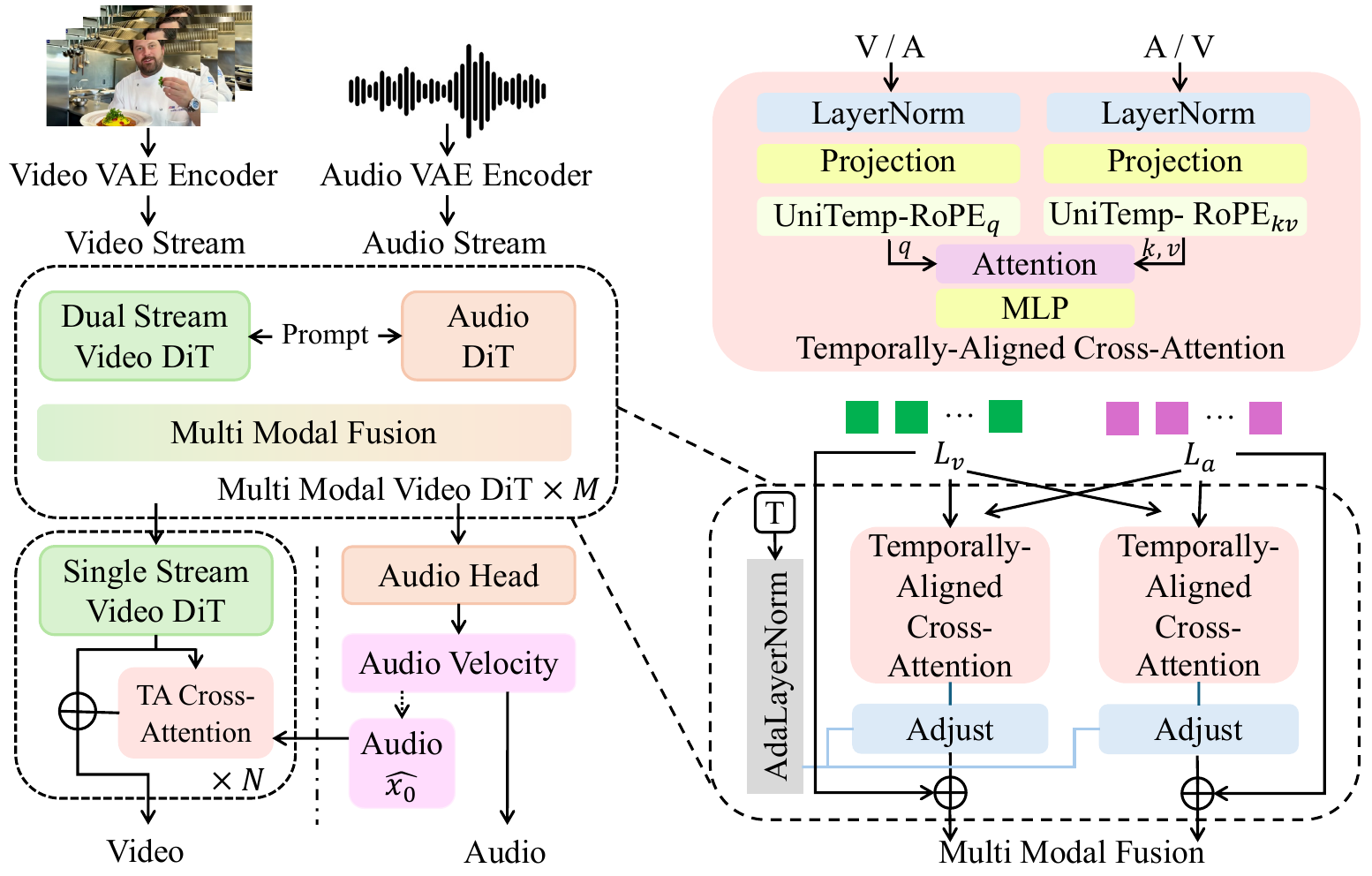}
    \caption{
        Architecture of \method.
    }
    \label{fig:dit}
\end{figure}

\subsection{Audio-Video Joint Modeling}

\paragraph{Prompt condition-controlled Audio-DiT.}

To enable efficient audio modeling in a compact space, we first employ a WavVAE~\citep{jiang2025megatts} to compress the high-dimensional raw audio waveform into continuous latent representations. This allows our model to focus on learning the semantic and acoustic structures rather than low-level waveform details.

\begin{figure}[htbp]
    \centering
    \includegraphics[width=1.0\textwidth]{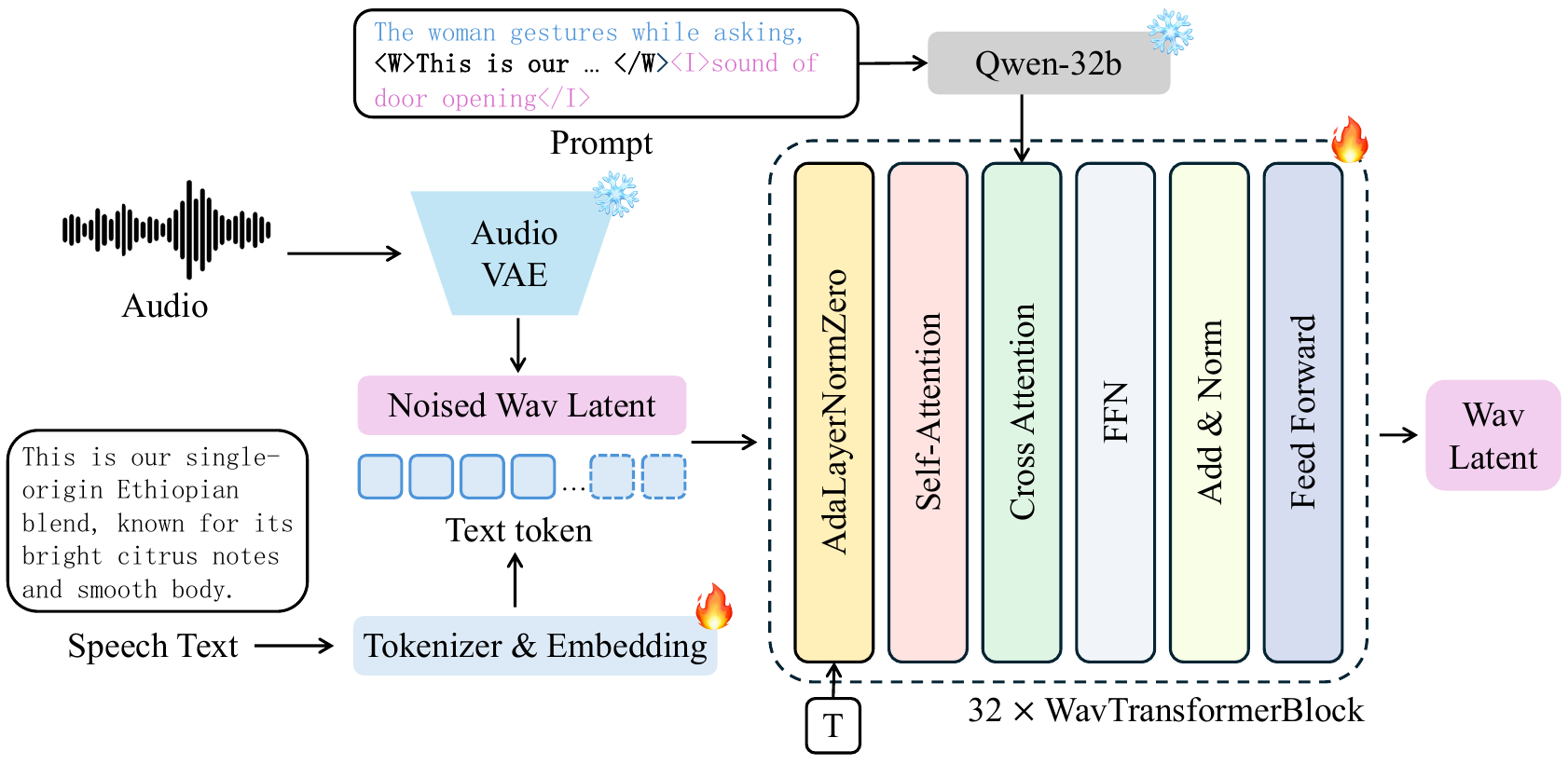}
    \caption{
        Architecture of Audio DiT.
    }
    \label{fig:audio_dit}
\end{figure}

As illustrated in Figure~\ref{fig:audio_dit}, the backbone of our audio branch is a Diffusion Transformer (DiT) consisting of 32 transformer blocks. Each block incorporates AdaLayerNormZero for timestep conditioning, self-attention for context modeling, and cross-attention for semantic guidance. To achieve precise control over both spoken content and acoustic atmosphere, we design a dual-conditioning mechanism that processes speech transcripts and descriptive prompts separately.

To ensure the generated audio strictly adheres to the desired linguistic content (i.e., what is spoken), we introduce the Speech Text as a direct input condition. Specifically, the target transcript is processed by a tokenizer~\citep{yang2025qwen3} and an embedding layer to yield a sequence of text tokens. These tokens are then aligned and concatenated with the noised audio latents to form the joint input sequence for the Audio DiT. This design forces the model to attend to the phonemic information directly at the input level, ensuring high intelligibility and content accuracy.

While the speech text dictates the content, the acoustic style (e.g., timbre, emotion) and background sounds are controlled by a global caption. We leverage the powerful Qwen-2.5-32B~\citep{qwen2.5} as our text encoder. As shown in the caption template (Figure~\ref{fig:audio_dit}, top), we utilize structured tags (e.g., \textless{}W\textgreater{} for transcripts, \textless{}I\textgreater{} for audio events) to disentangle the speech content from the acoustic description, while the overall prompt is described in a natural language manner. These tags are considered special tokens in the caption tokenizer. The resulting high-level semantic embeddings from Qwen are injected into the network via the Cross-Attention layers within each WavTransformer block. This allows the model to dynamically modulate the audio generation process, synthesizing specific sound events (e.g., ``sound of door opening") or speaker attributes guided by the rich semantic understanding of the LLM.

\paragraph{Unified Temporal Rotary Positional Embedding.}

Standard Rotary Positional Embeddings (RoPE)~\citep{su2024roformer} assume that the relative distance between tokens corresponds linearly to their indices (i.e., token $i$ and token $j$ have a distance of $i-j$). However, in our joint generation framework, the audio and video branches operate at significantly different sampling rates and latent temporal granularities. Let $L_a$ and $L_v$ denote the sequence lengths of audio and video latents, respectively. A direct index-based RoPE would erroneously assign the $t$-th video latent and the $t$-th audio latent to the same ``position," despite them representing vastly different physical timestamps.

To resolve this, we propose \textbf{UniTemp-RoPE}, which generalizes the discrete index domain of standard RoPE to a continuous temporal domain. We establish the audio latent sequence as the reference coordinate system.

Formally, for the audio sequence, the positional indices are naturally defined as integers:
\begin{equation}
    \mathbf{p}^{(a)} = \{0, 1, \dots, L_a - 1\}
\end{equation}
For the video sequence, instead of using discrete indices $\{0, \dots, L_v-1\}$, we map each video latent to its corresponding centroid timestamp in the audio coordinate sequence. Let $\Phi(\cdot)$ be the mapping function that projects the $i$-th video latent to the audio timeline based on the temporal receptive field center:
\begin{equation}
    \mathbf{p}^{(v)}_i = \Phi(i) \in \mathbb{R}, \quad \text{for } i \in \{0, \dots, L_v-1\}
\end{equation}
Here, $\mathbf{p}^{(v)}_i$ is a non-integer value representing the precise temporal center of the $i$-th video latent frame relative to the audio sampling rate. This mapping ensures that if a video frame temporally overlaps with a specific audio segment, their assigned positional values will be numerically close.

\begin{table}[!t]
\centering
\begin{tabular}{lccccccc}
\toprule
\textbf{Model} & \textbf{Model Size} & \textbf{M} & \textbf{N} & \textbf{Input Dim.} & \textbf{Output Dim.} & \textbf{Num of Heads} & \textbf{Head Dim} \\
\midrule %
VideoDiT       & 12B                 & 16         & 40         & 36                  & 16                  & 24                    & 128               \\
\midrule
AudioDiT       & 2B                  & 32         & --         & 32                  & 32                  & 24                    & 64                \\
\bottomrule
\end{tabular}
\caption{Key parameter selections in our Audio-Video DiT.}
\label{tab:dit_parameters}
\end{table}

\paragraph{Temporally-Aligned Cross-Attention.}
To enable precise interaction between the heterogeneous audio and video sequences, we propose the Temporally-Aligned Cross-Attention (TA-CrossAttn). Unlike standard cross-attention which relies on learned positional embeddings or relative biases that are agnostic to physical time, our mechanism explicitly incorporates the temporal alignment derived from UniTemp-RoPE.

A significant hurdle in joint generation is the distribution shift between the two modalities—audio latents (derived from spectral features) and video latents (derived from spatial-temporal patches) reside in distinct manifolds with different value ranges. Direct interaction can lead to optimization instability or one modality dominating the other. To address this, we utilize a Timestep-Aware Adaptive Layer Normalization before the fusion step. This module dynamically scales and shifts the incoming cross-attention output based on the current diffusion timestep $t$, ensuring that the injected cross-modal signals are normalized to a compatible range before being added to the residual stream.

Our architecture integrates TA-CrossAttn in a hierarchical manner, adapting to the specific requirements of different generation stages. 
In the dual stream blocks of VideoDiT, where the high-level semantic structure is established, we employ a symmetric bidirectional interaction. The VideoDiT queries the AudioDiT to align visual motion with sound onsets, while simultaneously, the AudioDiT queries the VideoDiT to match acoustic textures with visual scenes. This mutual communication ensures global semantic and temporal consistency. 
In the single stream VideoDiT, we shift to an asymmetric unidirectional interaction (Audio $\rightarrow$ Video). At this stage, the audio structure is largely determined. Compared to audio latents, video latents are more dense and diverse, and further synchronization enhancement is required. Unidirectional injection allows the video backbone to focus on high-fidelity visual synthesis while using the audio features as a strong, immutable condition, preventing late-stage visual fluctuations from corrupting the audio coherence.

\subsection{Audio-Video Refiner}
\label{sec:refiner}

Our base model is trained at 480p resolution. Directly training a joint audio-video model at 1080p to enable high-resolution audio-video generation would require substantial computational resources for both training and inference. To address this challenge, we further train an efficient cascaded audio-video (AV) refiner. The AV refiner is specifically designed to enhance the visual details while preserving the original video content and ensuring that neither audio quality nor audio-visual synchronization is degraded. Architecturally, the AV refiner builds upon the joint modeling approach described above and is initialized from the 480p audio-video base model.

\paragraph{Video Refinement}
On the video side, following \cite{zhang2025waver}, each 1080p training video is downsampled to 360p and then upscaled back to 1080p, simulating the low-resolution output of the first-stage generator. To further address generative artifacts and distortions from first-stage generator, we encode the upscaled video with a Video VAE to obtain latent representations, and inject timestep-dependent noise to produce noisy latents, which are then fed into the Video DiT of the AV refiner.

\paragraph{Audio Preservation}

To preserve audio quality and audio-visual synchronization, we systematically evaluated three architectural variants. Removing the Audio DiT and cross-attention modules and training an audio-driven video refiner forced the model to re-learn audio-visual synchronization, resulting in noticeably degraded lip-sync performance compared to the 480p base model. Alternatively, reusing the base AudioDiT model while injecting noise into the audio latents, in parallel with video refinement, maintained synchronization but introduced subtle and unpredictable changes to the audio, particularly in background sounds. In contrast, our adopted approach feeds clean audio latents into a frozen Audio DiT module during training, without adding noise, which effectively preserves the original audio fidelity and maintains the audio-visual synchronization established by the 480p base model.

%% file: content/3_data.tex
\section{Training Data}

\label{sec:data}
\begin{figure}[htbp] 
    \centering 
    \includegraphics[width=\textwidth]{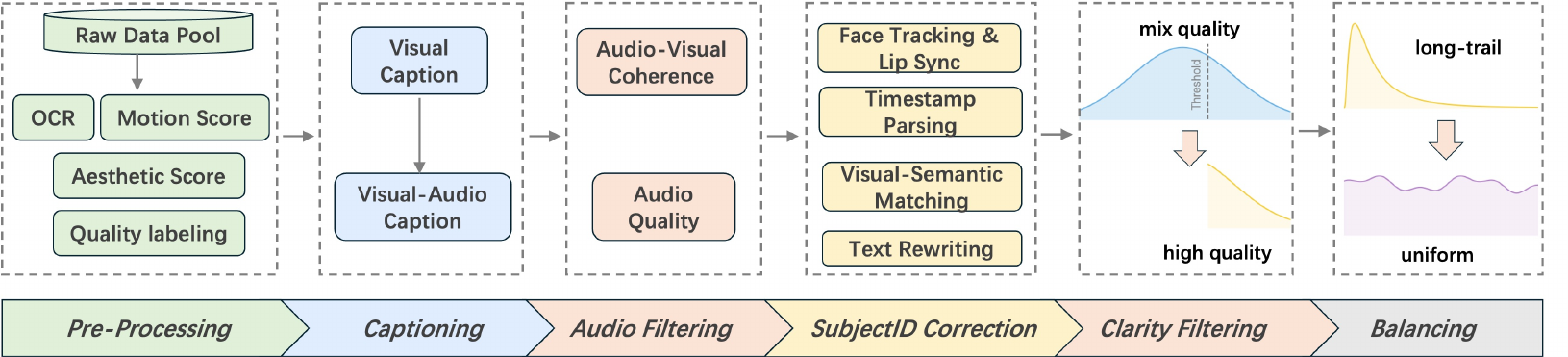} 
    \caption{An overview of our proposed data processing pipeline. The process consists of six main stages: video quality pre-processing, captioning, audio quality filtering, SubjectID correction, clarity filtering, and data balancing.}
    \label{fig:pipeline} 
\end{figure}

This section details the data processing pipeline for our joint audio-visual generation task. For such task, beyond ensuring standard video quality, we also address two additional critical dimensions: the intrinsic audio quality and the audio-visual coherence between the two modalities.
Our pipeline begins by filtering videos with audio from the raw data pool and then proceeds through six core stages: video quality pre-processing, captioning, audio quality filtering, SubjectID correction, clarity filtering, and data balancing. This process yields curated datasets tailored for various training phases. A flowchart illustrating these six stages is presented in Figure ~\ref{fig:pipeline}.

\subsection{Data Processing Pipeline}
\label{sec:data_process}
This section elaborates on the details of each stage in our data processing pipeline. The process commences by filtering videos that contain audio from the raw data pool. Subsequently, this data undergoes the following processing stages:

\textbf{Video Quality Pre-processing.} Building upon the video quality assessment methods from Waver1.0 ~\citep{zhang2025waver}, our process incorporates several key analyses. We employ Optical Character Recognition (OCR) to measure the proportion of overlaid text and detect watermarks; utilize a pre-trained aesthetics model for per-frame scoring; and compute optical flow using RAFT~\citep{2021RAFT} to analyze motion magnitude. Finally, a quality model (SFT on VideoLLaMA3~\citep{zhang2025videollama}) is used to classify samples as either `high-quality' or belonging to one of 13 distinct low-quality dimensions.

\textbf{Captioning.}
Our captioning process is a two-step procedure. First, we use our custom-trained visual caption model to generate visual contents, ensuring a consistent and controlled description of the visual elements. The visual description of each video is then fed into MLLM model ~\citep{comanici2025gemini}, along with the video, to incorporate the relevant audio information. (Further details are provided in Section ~\ref{sec:caption}).

\textbf{Audio Filtering.}
We utilize an MLLM model ~\citep{comanici2025gemini} to perform audio filtering based on two criteria. Firstly, for audio quality, the model assigns a score to each audio, allowing us to discard samples with significant background noise. Secondly, for audio-visual coherence, we assess the correlation and separate samples where the audio is highly correlated with the visual content, while also managing the proportion of weakly correlated data, such as background music (BGM), to optimize the dataset's composition.

\textbf{SubjectID Correction.}
In scenes involving people, we observe that the MLLM model ~\citep{comanici2025gemini} used for captioning is prone to errors in speaker-dialogue correspondence. These inaccuracies compromise downstream performance, particularly in speaking scenarios, by causing lip-sync animations to be applied to the wrong subject and degrading the model's ability to follow character-specific instructions. To mitigate this issue, we develop a correction pipeline built upon conventional computer vision models. (Further details are provided in Section ~\ref{sec:subjectid}).

\textbf{Clarity Filtering.}
To meet the demand for higher-resolution data for subsequent training stages, such as continuous training, SFT, and the 1080p refiner, we develop a specialized clarity filtering pipeline. We identify that conventional quality models have limitations in purely assessing clarity, as their judgments are often biased by irrelevant factors like color vibrancy or aesthetic composition.
To address this, our new pipeline employs an example-based scoring mechanism. It provides the evaluation model, e.g., ~\citep{comanici2025gemini}, with a set of reference images across six distinct clarity levels, from blurry to sharp. This forces the model to focus its evaluation on tangible texture details rather than subjective appeal. As a result, we can more objectively and accurately filter for high-definition videos rich in fine details, providing a cleaner dataset for our high-resolution model training.

\textbf{Data Balancing.}
To ensure the model learns concepts in a balanced manner, we perform strategic category balancing. First, we make a top-level distinction between core scenarios: `speaking' and `non-speaking' scenario. Then, specifically for data from general scenarios, we establish a comprehensive labeling system covering various domains such as animals, sports, transportation, and music (details in Section ~\ref{sec:data_label}).
Guided by prior knowledge, such as concept frequency and projected application scenarios, we then adjust the data proportions for each category. This approach ensures that high-value categories are sufficiently represented while preventing the model from being skewed by over-represented but less critical data, thereby achieving more effective and targeted learning.

\subsection{Caption Design}
\label{sec:caption}
\paragraph{Caption Schema.}

To facilitate precise control over the joint generation process and ensure semantic consistency across modalities, we design a structured caption schema $\mathcal{C}$. Unlike traditional unstructured captions, our schema disentangles the description into three distinct components: \textit{Subjects} ($\mathcal{S}$), \textit{Visual Context} ($\mathcal{V}$), and \textit{Interleaved Narration} ($\mathcal{N}$). The final input caption is constructed as a concatenated sequence,
\begin{equation}
    \mathcal{C} = \texttt{"Subjects: } \mathcal{S} \quad \texttt{Visual: } \mathcal{V} \quad \texttt{Narration: } \mathcal{N}\texttt{"}
\end{equation}
\vspace{-0.5em}
\noindent For instance, a constructed caption $\mathcal{C}$ would be instantiated as:
\begin{quote}
    \small
    \texttt{\textbf{Subjects:} Subject1: A man in a purple shirt with a clear voice. Subject2: A bearded man with a deep voice. \textbf{Visual:} A modern kitchen with grey cabinets. \textbf{Narration:} Subject1 points to Subject2 and says \textbf{<W>The vegetables are sold out</W>} accompanied by \textbf{<I>sizzling steak sounds</I>}. Subject2 turns around to cut vegetables.}
\end{quote}
\vspace{-0.5em}

Specifically, $\mathcal{S}$ serves as a registry of all salient entities (e.g., \texttt{Subject1}), defining both their visual appearance and, crucially, their \textbf{acoustic profiles} (e.g., ``deep, resonant voice"). $\mathcal{V}$ describes the static scene setting. $\mathcal{N}$ acts as the temporal backbone, chronicling the sequence of actions. Within $\mathcal{N}$, we introduce two sets of special tokens to anchor audio modalities: \textbf{\texttt{<W>}} encapsulates verbatim speech content (e.g., \texttt{<W>Hello</W>}), and \textbf{\texttt{<I>}} denotes non-speech acoustic events (e.g., \texttt{<I>footsteps</I>}). By leveraging the fluent, linear nature of natural language combined with these explicit tags, we inherently align auditory cues with visual descriptions in a unified narrative flow, ensuring that sound and sight are synchronized naturally through the text sequence.

\paragraph{Visual Caption model.} The caption provides an exhaustive account of the video content, covering aspects such as actions, entities, scene context, camera motion, stylistic attributes, spatial relationships, etc. In particular, action information is specified at a highly granular level. Such precise, fine-grained action annotations are critical for enhancing both motion realism/fidelity and instruction-following performance in video generation models, especially when adhering to detailed action directives.
\begin{itemize}
    \item \textbf{Supervised Fine-Tuning (SFT)} Caption SFT datasets were constructed using a Multimodal LLM. Subsequently, two types of manual annotations were performed based on these datasets: the first type consists of manually revised caption data, and the second involves decomposing sub-motion units to annotate the start and end timestamps of each complete small action described in the captions. Additionally, the Multimodal LLM was utilized to generate question-answer (QA) pairs regarding the positional relationships of subjects in the videos. The first round of SFT was conducted using three types of data: the constructed caption SFT dataset, the positional relationship QA pairs, and the sub-motion unit data. Subsequently, a second round of supervised fine-tuning (SFT) is conducted on the manually revised caption dataset.
    \item \textbf{Direct Preference Optimization~\citep{rafailov2024dpo} (DPO)} First, we leverage the fine-tuned model to perform 10 rounds of inference on a subset of the second-round SFT dataset. Subsequently, the Multimodal LLM is employed to quantify the occurrence frequencies of two distinct error types: motion description hallucinations and omissions of key action descriptions. Based on these error statistics, we construct preference pairs for Direct Preference Optimization (DPO) training: positive samples are defined as inference outputs free of both motion description hallucinations and key action omissions, while negative samples are selected as the outputs with the highest cumulative count of the two aforementioned errors across the 10 inference runs.
    Meanwhile, during the DPO training phase, we replace the standard DPO loss with DPOP~\citep{pal2024dpop} loss to mitigate the degradation issue commonly observed in long text generation.
\end{itemize}

\paragraph{Audio Event Annotation and Alignment.}
Building upon the detailed visual descriptions provided by the \textit{Visual Caption Model}, we employ a Multimodal LLM~\citep{comanici2025gemini} to inject acoustic information and enforce temporal alignment. Taking the raw video, audio, and the pre-generated visual caption as inputs, the annotation process focuses on three key dimensions:

\begin{itemize}
    \item \textbf{Acoustic-Visual Identity Binding:} We consolidate the visual attributes from the input caption and augment them with \textbf{vocal characteristics} (e.g., pitch, timbre) derived from the audio track. Each entity is assigned a unique ID (e.g., \texttt{Subject1}) to explicitly bind their visual appearance with their specific voice signature in $\mathcal{S}$.

    \item \textbf{Diegetic and Environmental Grounding:} To ground the scene acoustically, the narration $\mathcal{N}$ is initialized by defining the global acoustic atmosphere (e.g., "The overall environment sound is..."). We explicitly instruct the model to filter for \textbf{diegetic sounds}—those naturally originating from the visual scene—while discarding those irrelevant clips with background music.

    \item \textbf{Interleaved Event Extraction and Tagging:} We restructure the action sequence from the input visual caption to strictly align with the audio timeline. This involves extracting and inserting two types of acoustic signals directly into the narrative flow:
    \begin{enumerate}
        \item \textbf{Speech Transcription (\texttt{<W>}):} Verbatim dialogue is extracted and wrapped in \texttt{<W>} tags (e.g., \texttt{Subject1 says <W>Hello</W>}), ensuring the generated lip movements match the spoken content.
        \item \textbf{Acoustic Events (\texttt{<I>}):} Distinct sound effects are identified and marked with \texttt{<I>} tags (e.g., \texttt{<I>footsteps splashing</I>}) at the exact moment of occurrence, synchronizing the audio synthesis with the visual action.
    \end{enumerate}
\end{itemize}
By refining the visual captions with these acoustic details, we transform the text into a multimodal script where sight and sound are strictly aligned.

\subsection{Subject ID Correction Pipeline}
\label{sec:subjectid}
We have identified that the accuracy of SubjectID-to-Speech correspondence in captions is critical for generating high-quality results, particularly in multi-speaker scenarios. However, we observe that existing MLLMs, despite their advanced capabilities in comprehending audio and visual data, still exhibit limitations in consistently and accurately assigning SubjectID. Even in seemingly straightforward scenes with clear, front-facing, and non-occluded subjects, these models are prone to error, leading to incorrect speaker attribution.

\begin{figure}[htbp] 
    \centering 
    \includegraphics[width=1\textwidth]{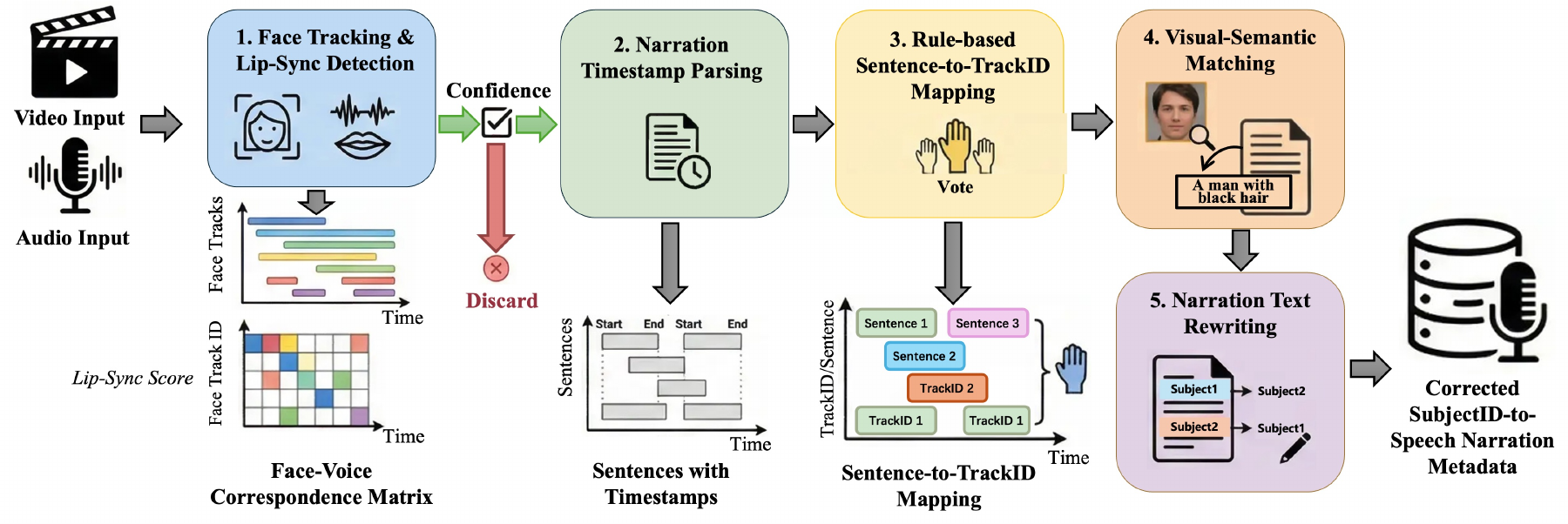} 
    \caption{Overview of our Subject ID correction pipeline, which aligns sentence-level speech segments with active face tracks via lip-sync detection and majority voting, then maps Track IDs to Subject IDs through visual-semantic matching and rewrites narration with corrected speaker labels.}
    \label{fig:correct_pipeline} 
\end{figure}
\vspace{-5pt}

To address this critical gap, we develop a more robust pipeline to correct SubjectID assignments in narration metadata. As shown in Figure~\ref{fig:correct_pipeline}, the pipeline consists of five main stages: tracking and lip-sync detection, narration parsing, sentence-to-trackID mapping, visual-semantic matching, and text rewriting, which collectively ensure a higher accuracy in SubjectID-to-Speech attribution.

\textbf{Step 1: Tracking and Lip-Sync Detection.}
We first process the video to extract the face track for each subject. And we also employ a detection model that jointly analyzes the audio and video streams to perform lip-sync detection. This process generates a face-voice correspondence matrix, which provides a matching score between each face track (TrackID) and the audio stream on a per-frame basis.
By applying a threshold to this matrix, we can determine which face is actively speaking at any given moment. This is represented as a binary matrix where the rows correspond to the number of TrackIDs and the columns correspond to the number of frames. A value of 1 in this matrix signifies active speech.

\textbf{Step 2: Narration Timestamp Parsing.}
Concurrently, we parse the original narration transcript using the Qwen3-omni~\citep{Qwen3-Omni} model. This step extracts each Speech Transcription along with its precise start and end timestamps. This temporal information is critical for aligning speech transcription with the speaker activity identified in Step 1.

\textbf{Step 3: Rule-Based Sentence-to-TrackID Mapping.}
With sentence timestamps and the active speaker matrix, we perform a rule-based Sentence-to-TrackID Mapping. For each sentence, we identify its corresponding time range in the active speaker matrix. The TrackID for that sentence is then assigned based on a majority vote: the face track with the highest number of active speaking frames within that time range is designated as the speaker. This process yields a preliminary TrackID for each speech sentence.

\textbf{Step 4: Visual-Semantic Matching and ID Assignment.}
This stage maps the TrackID to a permanent SubjectID. We use a multi-modal model to match the visual evidence of each TrackID (a face bounding box from a representative frame) against a set of predefined text subject descriptions, each TrackID linked to a unique SubjectID textual descriptions. By finding the best description-to-face match, the model establishes a definitive mapping from TrackID to SubjectID.

\textbf{Step 5: Narration Text Rewriting.}
In this stage, the corrected SubjectID mapping is used to rewrite the original narration text. The system replaces incorrect speaker labels in the narration with the newly assigned, accurate SubjectID. For example, a sentence originally attributed to Subject1 might be corrected to Subject2, ensuring the final output text accurately reflects the multi-modal analysis.

Throughout this pipeline, we implement a strict filtering strategy to ensure data quality. Videos that fail to meet a high confidence threshold at any stage, particularly during the lip-sync detection phase, are discarded. This curation process yields a high-quality dataset of clean examples.
A known trade-off of this approach is the potential for false negatives, where data featuring side-profiles or back-facing subjects may be inadvertently filtered out. However, we find that the resulting dataset, while simplified, is highly beneficial for the model during its initial training phase. It allows the model to effectively learn the fundamental task of subject-to-speech matching in straightforward scenarios.
Once the model has mastered these basic matching relationships, we can progressively enhance its adaptability by introducing more complex, manually verified data, such as examples with side-profiles and back-facing views.



\begin{figure}[t]
    \centering 
    \begin{subfigure}{0.95\textwidth} 
        \centering
        \includegraphics[width=\linewidth]{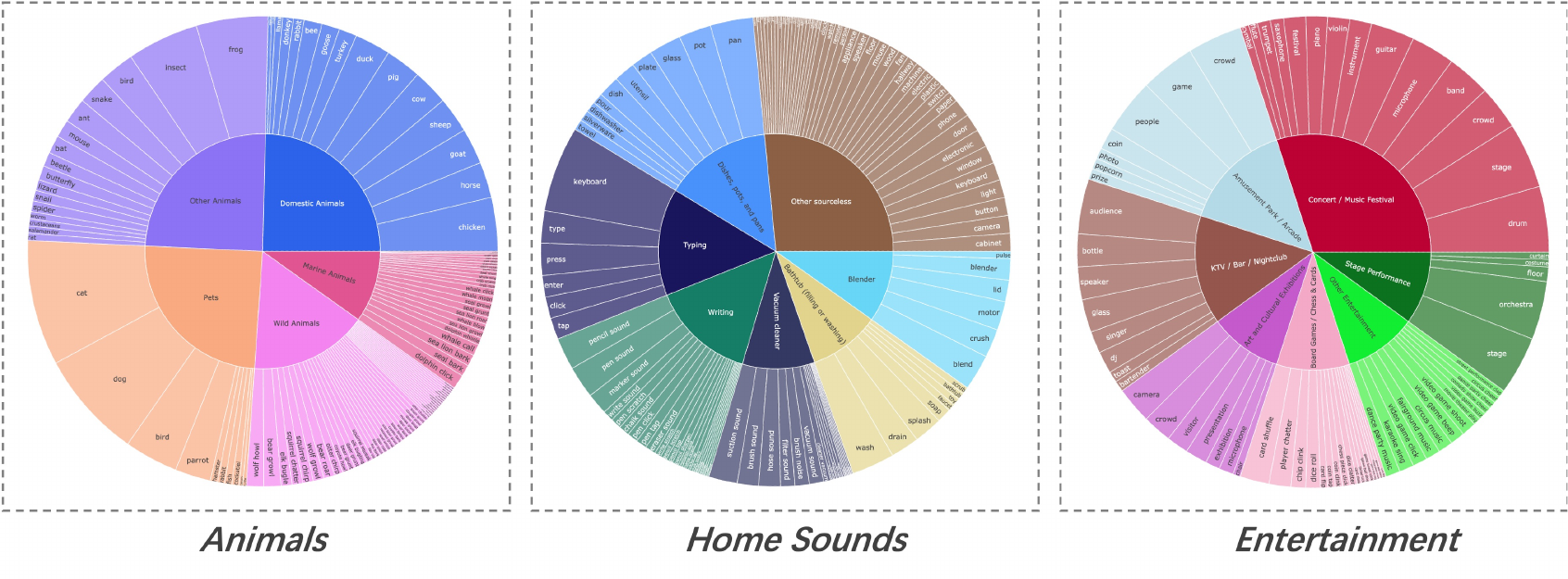}
    \end{subfigure}
    \hfill
    \begin{subfigure}{0.95\textwidth}
        \centering
        \includegraphics[width=\linewidth]{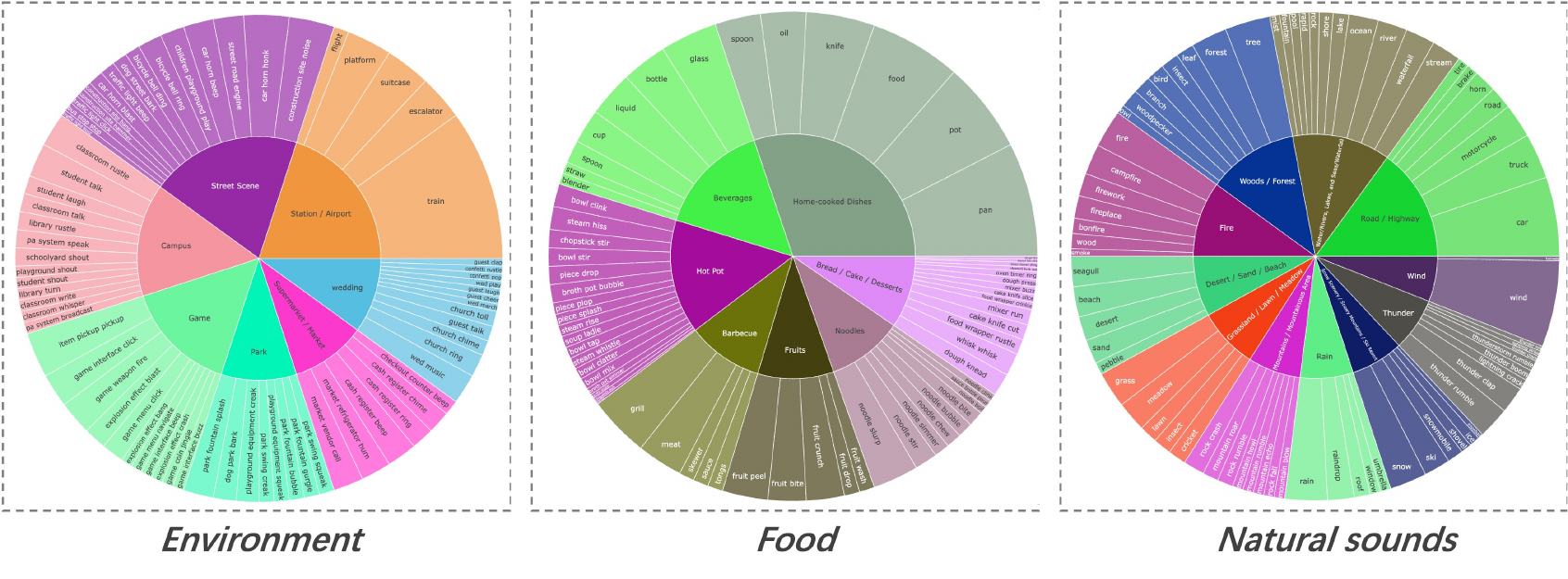}
    \end{subfigure}
    \hfill
    \begin{subfigure}{0.95\textwidth}
        \centering
        \includegraphics[width=\linewidth]{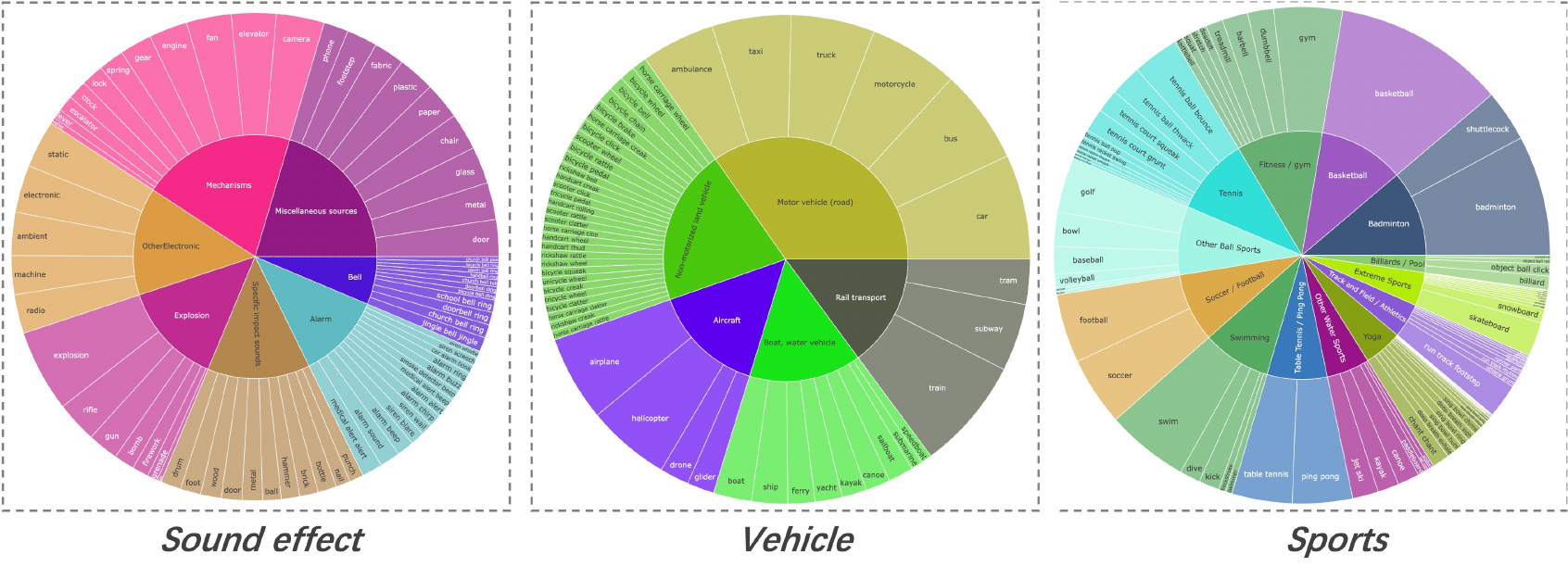}
    \end{subfigure}

    \caption{The detailed visual structure of the corresponding Level 2 and Level 3 sub-categories of nine Level 1 visual tags in General-Data Labeling System.}
    \label{fig:data-label}
\end{figure}
\vspace{-5pt}

\subsection{General-Data Labeling System}
\label{sec:data_label}
A fundamental challenge in audio-visual data curation is the one-to-many mapping between a single visual concept and its multiple potential audio events. For instance, the visual object `cat' can correspond to various sounds like `meowing' or `purring', while a `boat' might be associated with `water splashing', `horn blasts', or `engine sounds'.
To manage this complexity, our Labeling System is built upon a joint audio-visual keyword retrieval model. The process involves searching the captions of our entire dataset against a comprehensive, predefined list of keywords. Samples are tagged only when their captions match a specific audio-visual keyword pair.
This approach not only enables fine-grained categorization of our data for better organization and retrieval, but it also acts as an effective filtering mechanism to screen out ``silent" instances that lack explicit audio events (e.g., a cat making no sound). This increases the hit rate of audio signal generation in the final output.

Specifically, the system is organized into a three-level hierarchy. The top-level (Level 1) categories consist of nine main domains: Animals, Home Sounds, Entertainment, Environment, Food, Nature, Sound Effects, Vehicles, and Sports. 
Guided by prior knowledge, such as usage frequency and projected application scenarios, we strategically adjusted the data proportion for each category. The detailed structure of the corresponding Level 2 and Level 3 sub-categories is illustrated in Figure ~\ref{fig:data-label}.
Crucially, for each Level 3 visual tag, we then generate a corresponding list of associated audio keywords. This process establishes the final `visual tag + audio keyword' pairs that are used for our joint retrieval process.

\subsection{Character-Driven Data Pipeline}
To empower the model with robust identity-preserving capabilities for Sora-like role-playing animate, we integrated character-driven data into the supervised fine-tuning (SFT) phase. Specifically, we performed further refinement—including targeted filtering and processing—on the preprocessed human-specific dataset, and devised two specialized data pipelines to this end: the Cross-pair Data Pipeline for mining real-world temporal variations, and the In-pair Data Augment Pipeline for enhancing reference data diversity.

\textbf{Cross-pair Data Pipeline}
The Cross-pair pipeline focuses on identifying the same individual across different contexts and temporal segments within long-form videos. The pipeline can be divided into the following steps:

\begin{itemize}
    \item \textbf{Segment Selection and Audio-Visual Anchoring:} From raw source videos (10-30 minutes), we extract $N$ short clips of 3-10 seconds. To ensure the highest quality of audio-visual alignment, we utilize TS-TalkNet~\citep{jiang2023target} to evaluate the active speaker. For each detected subject, we select the 1.5-second sub-clip with the maximum sync score. This segment serves as the "Identity Anchor," containing the most representative facial features and corresponding vocal timbre. 

    \item \textbf{Multi-Subject Feature Matching:} Since a single clip may contain multiple individuals, we identify $M$ distinct subjects ($M > N$). We designate the middle frame of each segment as the representative frame $F_{mid}$. We employ a dual-feature approach: CLIP~\citep{radford2021learning} embeddings for capturing global visual semantics and ArcFace~\citep{deng2019arcface} embeddings for fine-grained facial identity verification. Then we construct an $M \times M$ similarity matrix using these features to map subjects across different clips.

    \item \textbf{Cross-pair Subject Selection:} In order to enable the model to learn motion representation rather than conduct static replication, we select image pairs depicting the same individual in distinct states and impose three criteria for valid cross-pairs, namely a face similarity score computed via the ArcFace algorithm greater than 0.35, a global similarity score derived from the CLIP model higher than 0.7, and a proximity criterion requiring the absolute similarity score to be closely proximate to 0.9, which effectively prevents the model from being trained on quasi-identical frames (a major cause of copy-paste artifacts) while ensuring the unambiguous consistency of the target subject’s identity.
  
\end{itemize}

\textbf{In-pair Data Augment Pipeline} To mitigate the pervasive ``copy-paste" bias—where the model trivially replicates pixels from the reference image rather than learning latent motion dynamics—we implement a comprehensive Reference Augmentation Pipeline for in-pair reference image. Our augmentation process comprises the following specialized operations: pose and expression perturbation, contextual decoupling via background augmentation and semantic editing. 

Previous works used ControlNet~\citep{zhao2023uni,zhang2023adding} to retarget body skeletons for diverse poses and facial expressions, aiming to enable the model to animate subjects across varied expressions and poses. Additionally,  ICLight~\citep{zhang2025scaling} has been adopted for physics-aware relighting and background illumination adjustment. Furthermore, they combine Segment Anything Model (SAM)~\citep{kirillov2023segment} for precise subject segmentation and Flux-Inpainting~\citep{labs2025flux} for background replacement, to separate identity features from background and illumination cues. In our work, our comparison reveals that the current state-of-the-art (SOTA) unified editing model performs better on the aforementioned editing tasks. Ultimately, we adopted Qwen-Image-Edit~\citep{wu2025qwenimagetechnicalreport} to execute the above-mentioned data augmentation tasks.

By training on these ``augmented-reference vs. original-target" pairs, \method is compelled to prioritize identity-level conditioning (e.g., ``who the person is") over pixel-level replication (e.g., ``what the image looks like"), leading to significantly improved motion amplitude and character consistency in complex scenarios.

\subsection{Hierarchical Data Filtering}
\label{sec:data_Hierarchical}
We employ a hierarchical data filtering strategy across the entire training pipeline, spanning from 480p joint-training and continuous training to SFT and the final 1080p Refiner stage. The core idea is to shift the learning focus as the model matured:
In early stages (Joint-training / Continuous Training): The initial focus is on the acquisition of broad knowledge and establishing foundational visual quality, aiming for the model to grasp a wide range of real-world concepts.
In later stages (SFT / Refiner): As the model's foundational capabilities solidified, the emphasis gradually shifts towards refining specific attributes, such as improving prompt fidelity, enhancing visual aesthetics, and elevating the joint audio-visual quality.

\begin{figure}[htbp]
    \centering
    \includegraphics[width=0.6\textwidth]{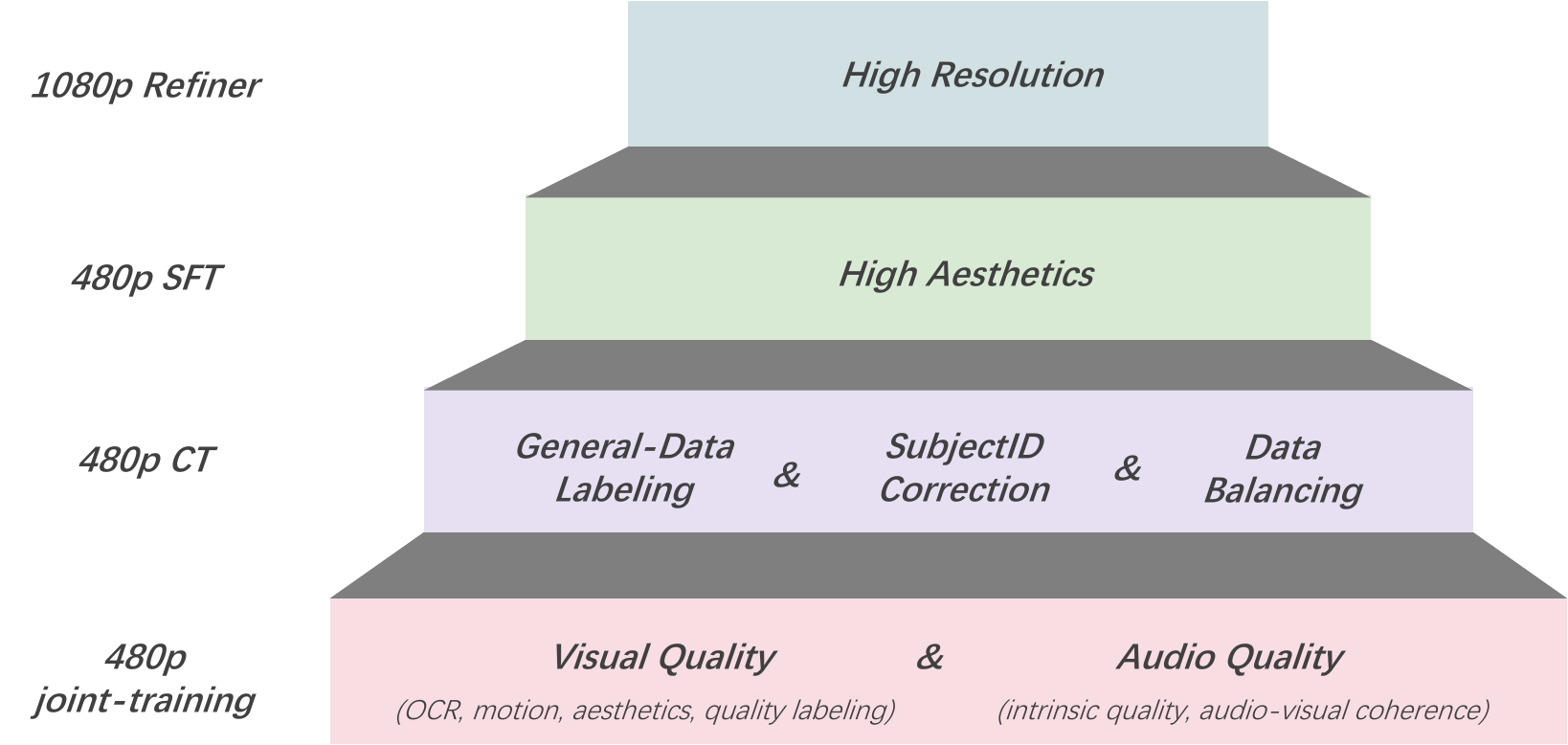}
    \caption{
        The hierarchical data filtering funnel, which progressively refines the dataset through increasingly strict quality criteria.
    }
    \label{fig:data-filter}
\end{figure}

\textbf{480p Joint-Training.}
The goal of data construction at this stage is to build a large-scale, high-quality baseline dataset. We focus on comprehensive filtering for both visual quality (covering OCR, motion analysis, aesthetics, and quality model labeling) and audio quality (based on intrinsic quality and audio-visual coherence).

\textbf{480p Continue-Training (CT).}
In this stage, the focus shifts to improving the balance and accuracy of prompt following. We perform data balancing for speaking versus non-speaking scenarios and for general scene categories, and also rectify SubjectIDs in multi-person dialogue data to enhance character-matching accuracy.

\textbf{480p Supervised Fine-Tuning (SFT).}
This stage is dedicated to enhancing the model's aesthetic performance. We introduce a significant portion of high-aesthetics data, mixing it with realistic data at an approximate 3:1 ratio, and filter out blurry samples based on their clarity scores.

\textbf{1080p Refiner.}
Tailored for the 1080p high-resolution refiner, this stage employs strict data selection criteria. We apply a high threshold to the clarity scores to curate a dataset of extremely high-definition samples rich in detail, while also refining the data distribution based on general-data labels to meet the specific demands of high-resolution generation.

%% file: content/4_recipe.tex
\section{Training / Inference Recipe}

In this section, we elaborate on our multi-stage training pipeline and inference strategies. We specifically discuss the disparity in training dynamics between audio and video modalities during joint generation training, and further introduce the techniques to elevate visual aesthetics, color fidelity, audio-visual synchronization and character expressiveness of the generated output.

\begin{table}[!t]
\centering
\renewcommand{\arraystretch}{1.3} 
\small
\resizebox{\textwidth}{!}{
\begin{tabular}{l|cccccc}
\shline
\multirow{2}{*}{\textbf{Task}} & \multirow{2}{*}{\textbf{Stage}} & \multirow{2}{*}{\textbf{Resolution}} & \multirow{2}{*}{\textbf{Data volume}} & \multirow{2}{*}{\textbf{Epochs}} & \multicolumn{2}{c}{\textbf{Learning rate}}             \\ \cline{6-7} 
                               &                                 &                                      &                                       &                                  & VideoDiT               & AudioDiT / CA in V2A      \\ \shline
T2A                            & Stage I                        & -                                    &    384M                                   &        1                          & -                      & \multicolumn{1}{c}{5e-5} \\ 
T2A                            & Stage II                        & -                                    &    19M                                   &     10                             & -                      & \multicolumn{1}{c}{5e-5} \\ \shline
T2VA                           & Joint-Train                        & 480p 24fps                           & 11M                                   & 1.2                              & 1e-4                   & 1e-4                      \\
T2VA+I2VA                      & Joint-Train                        & 480p 24fps                           & 11M                                   & 0.3                              & 5e-5                   & 5e-5                      \\
T2VA+I2VA                      & CT                              & 480p 24fps                           & 4.3M                                  & 3                                & 5e-5                   & 5e-5                      \\
T2VA+I2VA                      & SFT                             & 480p 24fps                           & 5M                                    & 0.5                              & 1e-5                   & 1e-6                      \\ \shline
(T+V)A2VA                      & Refiner                          & 1080p 24fps                          & 0.7M                                  & 1                                & 5e-5                   & 5e-5                      \\ \shline
\end{tabular}
}
\caption{The entire training process for the four tasks: T2A, T2VA, I2VA, and Refiner. ``M'' refers to million. ``CA in V2A" denotes audio-related
Video-to-Audio cross-attention modules.}
\label{tab:training_stages}
\end{table}

\subsection{Multi-stage Training}
Our entire multi-stage training schedule is shown in Tab.~\ref{tab:training_stages}, which includes T2A, T2VA, I2VA, and Refiner tasks. We also show the data volume, number of training epochs, and learning rate strategy for each stage.

\textbf{T2V.}
We initialize the VideoDiT backbone directly using the pre-trained weights from Waver1.0-480p-SFT version ~\citep{zhang2025waver}.

\textbf{T2A.} 
The fidelity of the generated audio is fundamentally bounded by the reconstruction capability of the VAE. We adopt the pre-trained VAE from MegaTTS3~\citep{jiang2025megatts} as our initialization. However, as the original model was tailored specifically for Text-to-Speech (TTS) tasks, it lacks the capacity to handle the complex spectral diversity found in general video audio (e.g., background music, environmental noise). To bridge this domain gap, we perform continuous training on the VAE. We construct a composite dataset merging the original TTS data with high-fidelity music tracks, sound effect libraries, and large-scale in-the-wild video audio. This adaptation ensures that our latent space is robust enough to reconstruct diverse acoustic phenomena beyond human speech.

Training the Audio DiT presents a unique challenge: speech generation requires high precision and low entropy (i.e., strict adherence to linguistic content), whereas environmental sound generation allows for higher degrees of freedom. To balance these conflicting objectives, we design a two-stage curriculum learning strategy:

\begin{itemize}
    \item \textbf{Stage I: Large-scale TTS Pre-training.} We prioritize the learning of linguistic-acoustic alignment. The model is trained exclusively as a TTS system using a massive corpus of \textbf{700k hours} of transcribed speech. All audio is sampled at 16kHz. This stage forces the model to master text-to-spectrogram conversion and capture diverse speaker timbres, establishing a strong foundation for speech intelligibility.
    
    \item \textbf{Stage II: Multi-Condition Instruction Tuning.} Building upon the pre-trained TTS weights, we transition to the joint generation task. In this stage, the model is conditioned on both speech transcripts and descriptive prompts (captions). We curate a mixed dataset comprising a high-quality subset of the pure speech data (approx. \textbf{5k hours}) and a large-scale video dataset with dense sound event annotations (\textbf{111k hours}). This fine-tuning stage enables the model to generalize from pure speech synthesis to complex, prompt-controlled audio-visual scene generation, seamlessly blending dialogue with background dynamics.
\end{itemize}

\textbf{T2VA \& I2VA \& R2VA. } Upon initializing AudioDiT and VideoDiT weights, we conduct full-parameter training across the AudioDiT, VideoDiT, and cross-attention layers using 480p/24fps video samples. The multi-stage pipeline is structured as follows: 1) Joint-training: We initially train on an 11M dataset for the T2VA task. After the first epoch, the I2VA task is introduced with a probability of 0.4. 2) Continue Training (CT): We then train on a class-balanced dataset of 4.3M samples for approximately 3 epochs. 3) Supervised Finetuning (SFT): We incorporate high-aesthetic data to specifically elevate visual attributes, aiming for cinematic lighting and texture. Considering that the audio modality tends to converge rapidly on limited data, we maintain a data scale comparable to the previous stage rather than performing a significant reduction, to ensure robust audio (detailed in Sec ~\ref{sec:avdynamics}). 4) Reference Joint Tuning (RJT): To enable character-driven video generation for role-playing scenarios, we integrate reference-image guidance capabilities into the existing framework. Leveraging a curated dataset of 0.8M reference-paired samples, we jointly train this stage with high-quality SFT data. Furthermore, we carefully design the conditional guidance mechanism for reference images, which enhances character fidelity without compromising the model’s original T2VA and I2VA capabilities (detailed in Sec ~\ref{sec:roleplaying}).

\textbf{Training hyperparameters. }
We set the sigma shift ~\citep{esser2024sd3} in flow matching to 7. For Learning rate, in text-to-audio training, we utilize a learning rate of $5e\!-\!5$ for both two-stage training. During joint generation, we initialize at $1e\!-\!4$ for T2VA joint-training, then reducing it to $5e\!-\!5$. In the SFT stage, we employ small, asymmetric learning rates for the Video, Audio, and Video-to-Audio cross-attention. This strategy enables updates to the visual distribution while preserving the stability of audio signal weights.

\subsection{Disparate Audio-Visual Training Dynamics}
\label{sec:avdynamics}

In joint training, we observe that the Audio and Video modalities exhibit different training dynamics. This divergence likely stems from disparities in model size and latent scale between the two branches, leading to inconsistent learning behaviors and convergence speeds under identical hyperparameter settings. Consequently, strategic trade-offs must be made to balance the learning progress of both modalities across different training stages.

\textbf{The Importance of Audio Training.}
During the initial joint-training phase, the effective data volume for the joint distribution is limited compared to large-scale audio-only datasets. As a result, it is difficult to fundamentally alter the base quality of the audio during this stage.
We observe that the performance of the initial  AudioDiT training (T2A), e.g., tone authenticity, pronunciation accuracy, emotional consistency, essentially determines the performance upper bound of the audio modality in joint generation. 
In the subsequent joint training phase, the model tends to learn audio-visual synchronization. While we observe minor adaptations in environmental sounds, base audio quality such as tone authenticity and pronunciation accuracy do not significantly improve.
Therefore, if the base audio training is deficient, attempting to rectify it through continued joint training is ineffective.

\textbf{Audio Sensitivity and Forgetting.}
We observe that the audio modality exhibits heightened sensitivity to distributional shifts, a problem that becomes acute when fine-tuning on high-quality SFT data after large-scale training and data balancing.
When transitioning to SFT stage with shifting data distributions (e.g., adjustments in data ratio or scale), the audio branch tends to converge rapidly to the new distribution. This often results in catastrophic forgetting of robust knowledge acquired from earlier large-scale datasets.
To address this rapid degradation of audio performance under SFT-driven distribution shifts, we experiment with three targeted SFT strategies on the high-quality data, and Asymmetric Learning Rates achieves optimal performance:

(1) Freezing AudioDiT (Rigid Constraint):
We attempt to freeze the weights of the converged AudioDiT, relying solely on the cross-attention modules to adapt to the new data distribution for synchronization. However, audio generation relies on the tight coupling between the AudioDiT backbone and the cross-attention mechanism. 
This approach proves too rigid. By fixing the backbone, we force the cross-attention layers to compensate excessively for distribution shifts. This ``misalignment" between the frozen audio features and the adapting visual context frequently corrupted the generated audio, introducing audio artifacts such as high-frequency distortion and electronic noise. Thus, we suppose that AudioDiT and cross-attention must work in coordination, although they may adapt at different speeds.

(2) Asymmetric Learning Rates (Optimal Strategy):
To enable collaborative optimization while protecting audio stability, we propose a differential learning rate strategy. We assign a larger learning rate to the VideoDiT backbone to facilitate aggressive updates of the new visual distribution. Conversely, we set a significantly smaller learning rate for the AudioDiT and the audio-related Video-to-Audio cross-attention modules.
This strategy proves to be more robust. It allows the video model to undergo necessary changes for high-quality generation, while the audio model maintains its stability, making only micro-adjustments for synchronization. This effectively prevents the audio weights from drifting into the new or smaller-scale data distribution, preserving original audio quality.

(3) Model Fusion: We explore training a separate, independent SFT model on the new data and then merging it into the base model with a low weight ($\approx 0.1$). The goal is to inject domain-specific knowledge without disrupting the base model's stability.
While the independent model learns the new domain well, it suffers from catastrophic forgetting regarding the general domain knowledge.
Even with a low fusion weight, the corrupted parameters from the specialized model interfered with the base model's general capabilities.
This negative interference appears as occasional audio distortion and instability in non-target domains. It shows that a simple weighted fusion of model parameters cannot easily resolve the conflict between specializing on new data and maintaining the general robustness of the base model.

\subsection{Aesthetics Optimization}

In the aesthetics optimization stage, the primary objective is to enhance the aesthetic quality of the videos, guiding the model to generate visuals with cinematic-level lighting and composition. This is achieved while ensuring the original audio quality is maintained without degradation.
Below, we introduce our specific methods from the perspectives of both data and training.

\begin{figure}[ht]
    \centering
    \includegraphics[width=\textwidth, trim=0 160 0 160, clip]{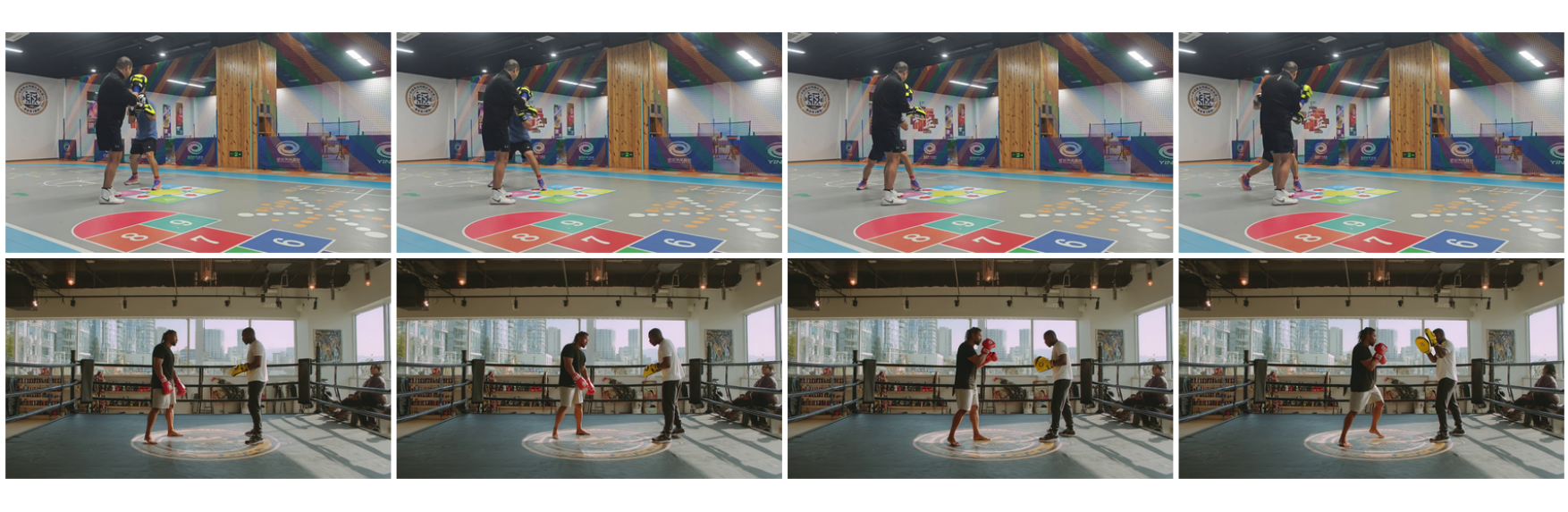}
    \caption{
        Aesthetic comparison for a boxing scene. \textbf{Top}: Typical real-world video. \textbf{Bottom}: Our synthetic data shows superior visual quality, including better lighting, color and composition.
    }
    \label{fig:data_comparison}
\end{figure}

\textbf{Aesthetic Data Acquisition.}
Leveraging our unified multi-task model, we curate a diverse set of synthetic video samples by combining high-quality image corpora with synthesized surreal imagery and our I2V generation capability. The resulting videos are deliberately produced with an emphasis on aesthetic fidelity—e.g., visually compelling content, refined composition, and nuanced color and lighting. The strategy  also improves the model's capacity for creative instruction following and its robustness to surreal, non-photorealistic concepts. As illustrated in Figure~\ref{fig:data_comparison}, our synthetic data exhibits a significant improvement in aesthetic quality over typical real-world footage.

\textbf{Aesthetics Finetuning.}
We conduct finetuning using this curated high-aesthetic data.
Given that this data exhibits large visual variance but a narrow audio distribution, and considering the modality discrepancies described in Sec. ~\ref{sec:avdynamics}, we employ the Asymmetric Learning Rates strategy.
Specifically, we set a learning rate of 1e-5 for VideoDiT to update the visual style, and a 1e-6 learning rate for AudioDiT and the Video-to-Audio cross-attention layers. For the training data, acknowledging the instability of audio on small-scale datasets, we slightly downsample the data from the continued training (CT) stage and mix it with the high-aesthetic data at a 3:1 ratio. This approach allows us to refine the visual style while preserving audio robustness.
Furthermore, we observe during inference that using prompts stylistically similar to the high-aesthetic data makes it easier for the model to leverage the learned high-aesthetic distribution, preventing it from reverting to the original, lower-quality visual patterns.

\begin{figure}[!t]
    \centering
    \includegraphics[width=1.0\textwidth]{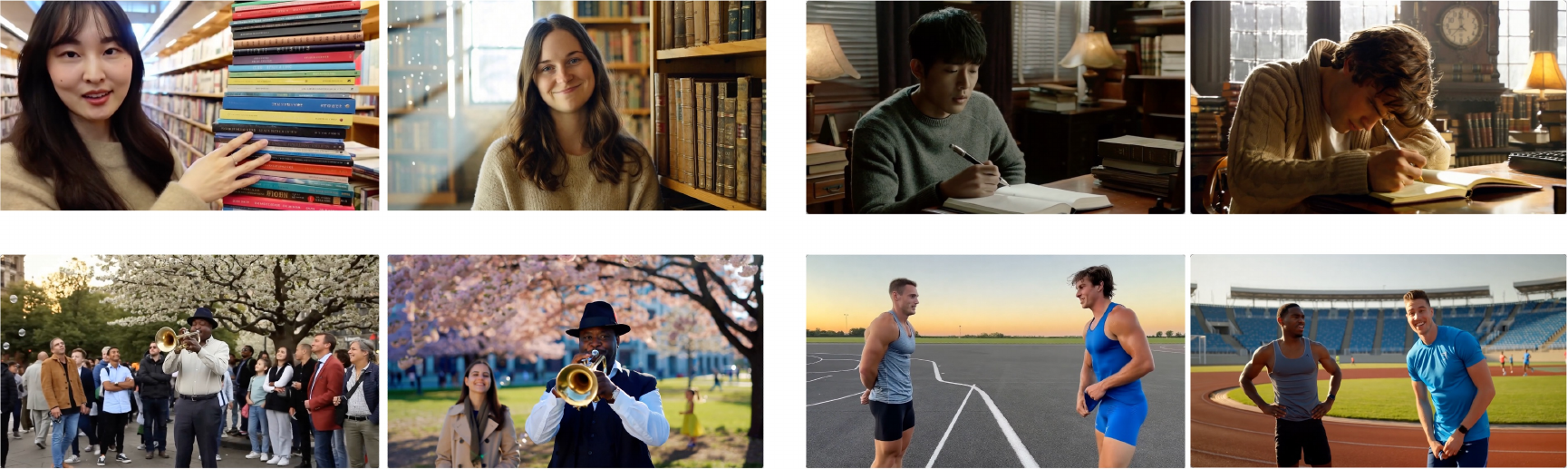}
    \caption{
        Comparison of four videos generated by models before and after high-aesthetic finetuning. For each case, the frame on the \textbf{left} is before finetuning, and that on the \textbf{right} is after finetuning. 
    }
    \label{fig:aes_show}
\end{figure}

This fine-tuning has brought about a significant improvement in visual aesthetics, color, and clarity, which we will demonstrate below from both quantitative and qualitative perspectives.
 \textbf{Quantitative improvements} are demonstrated in a side-by-side human evaluation on a 264-prompt general benchmark (Alive-bench 1.0). The fine-tuned model achieved a \textbf{25.41\% win rate} in visual quality, substantially outperforming the base model's 10.65\%. Crucially, the evaluation also confirms that audio
 quality remains comparable between the two models, indicating our approach successfully isolates and enhances aesthetic performance.
 \textbf{Qualitative enhancements} are clearly visible in generated samples (Figure~\ref{fig:aes_show}). After fine-tuning, videos exhibit noticeable improvements in color vibrancy, brightness, clarity, and overall compositional harmony.

\subsection{Inference Recipe}
During the inference stage, we employ specific strategies aimed at improving the final visual quality and enhancing the injection of audio-visual synchronization information.

\textbf{Multi-Condition Controlled Generation via CFG.}
In our joint generation framework, the model is guided by two distinct conditions: the text prompt, which dictates the semantic content, and the cross-attention signal, which ensures audio-visual synchronization by allowing each modality to receive information from the other.
Initially, without this cross-attention mechanism, both the Audio and Video branches can generate their respective outputs independently. The introduction of cross-attention provides a mutual signal that steers the model toward audio-video synchronization, effectively establishing a secondary condition that specifically governs synchrony.
Based on this, we adopt a multi-condition control scheme, treating the text prompt (positive $c_{pos}$ / negative $c_{neg}$) and the mutual cross-attention signal ($c_{mutual}$) as distinct, controllable conditions for guidance.
Thus, we can divide the state into four types based on the presence or absence of each condition: $\epsilon_{{pos,mutual}} = \epsilon_\theta(x_t, c_{{pos}}, c_{{mutual}})$, $\epsilon_{{pos,indep}} = \epsilon_\theta(x_t, c_{{pos}}, \emptyset_{{mutual}})$, $\epsilon_{{neg,mutual}} = \epsilon_\theta(x_t, c_{{neg}}, c_{{mutual}})$, $\epsilon_{{neg,indep}} = \epsilon_\theta(x_t, c_{{neg}}, \emptyset_{{mutual}})$.
During training, we implement a dropout strategy to ensure the model performs well in each state. With a probability of 0.3, we drop the $c_{mutual}$ component, which deactivates the cross-attention information exchange between the AudioDiT and VideoDiT. This forces each model to learn independently, preserving its ability to generate high-quality outputs even without cross-modal input.

\begin{figure}[t]
    \centering
    \includegraphics[width=1.0\textwidth]{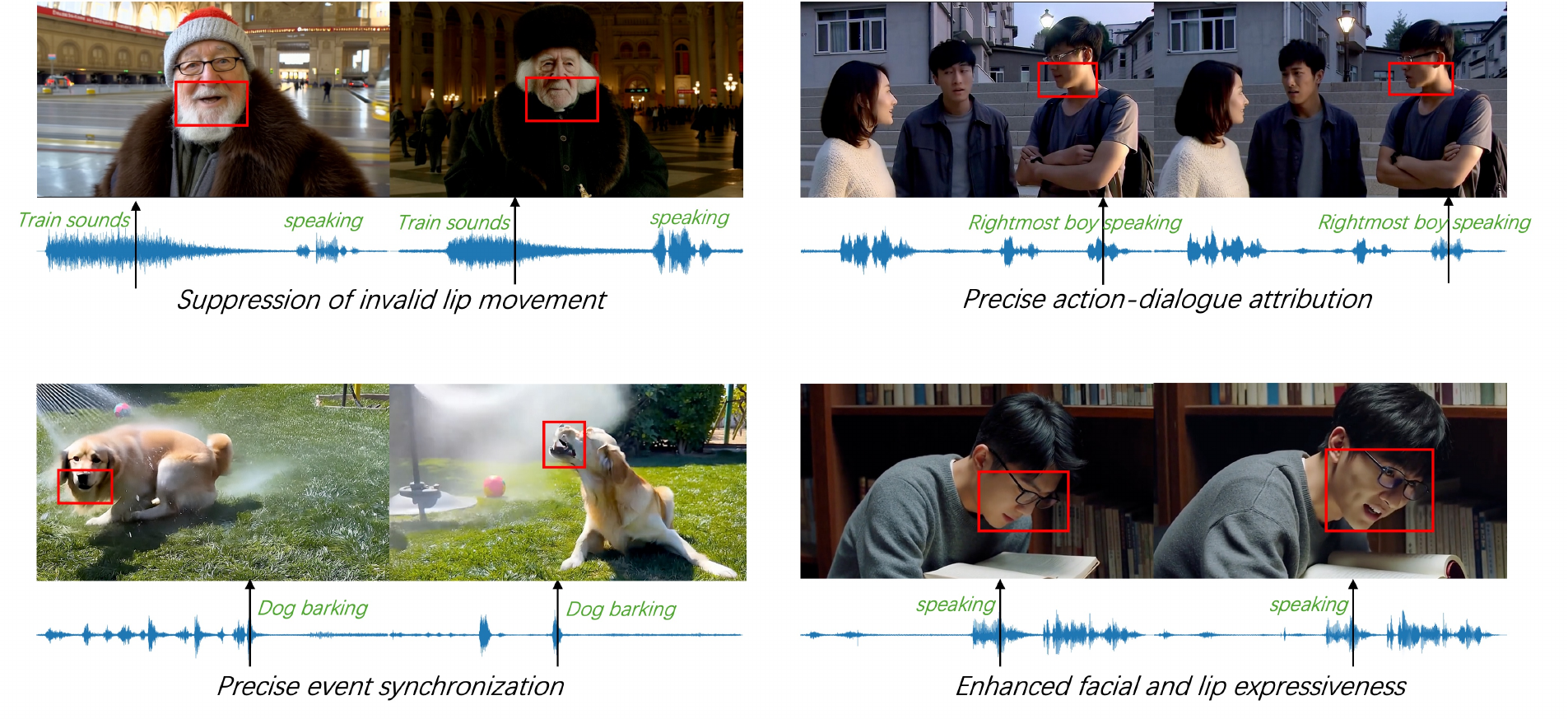}
    \caption{
        Comparisons of results generated using only the text condition (\textbf{left}) versus using both text and the mutual cross-attention signal condition (\textbf{right}).
        The blue image below is the video's audio waveform. On it, green text labels the audio event, while the black arrow marks the current frame's position.
    }
    \label{fig:sharecfg}
\end{figure}

Our experiments show that incorporating CFG for the shared signal $c_{mutual}$ strengthens the synchronization of audio-visual events and improves character expressions and lip movements. Figure ~\ref{fig:sharecfg} contrasts the visual results generated using only the text condition versus using both text and the mutual cross-attention signal condition.

\textbf{Multi-Condition APG.}
Building upon Waver1.0 ~\citep{zhang2025waver}, we utilize Asymmetric Perturbation Guidance (APG) ~\citep{sadat2024eliminating} to enhance the realism of video and audio, improve visual color saturation, and mitigate artifacts. Since the original APG is designed for single-condition control, we adapt it to our multi-condition framework using a composite formula that balances the influence of each condition.
Within each term of this formula, we preserve the core APG mechanism: the guidance vector (i.e., the update difference) is decomposed into components parallel and perpendicular to a reference prediction, and the magnitude of the parallel component $eta$ is selectively dampened.
Specifically, we consider three different guidance vectors:
$\left(\epsilon_{{pos,mutual}}-\epsilon_{{neg,indep}}\right)$, $\left(\epsilon_{{pos,mutual}}-\epsilon_{{neg,mutual}}\right)$, $\left(\epsilon_{{pos,mutual}}-\epsilon_{{pos,indep}}\right)$.
Let $w_1, w_2, w_3$ represent the weights of each guidance vector, respectively. Let $\mathcal{A}$ represent the APG operation. Then, the final guidance noise $\hat{\epsilon}_\theta$ is computed as a weighted sum:
\begin{align}
    \Delta_{total} &= \mathcal{A}(\epsilon_{pos, mutual} - \epsilon_{neg, indep}) \\
    \Delta_{text} &= \mathcal{A}(\epsilon_{pos, mutual} - \epsilon_{neg, mutual}) \\
    \Delta_{mutual} &= \mathcal{A}(\epsilon_{pos, mutual} - \epsilon_{pos, indep})
\end{align}
\begin{equation}
    \hat{\epsilon}_\theta = \epsilon_{neg, indep} + w_1 \Delta_{total} + w_2 \Delta_{text} + w_3 \Delta_{mutual}
    \label{eq:final_guidance_no_text}
\end{equation}
We use the APG method in both VideoDiT and AudioDiT, but the parameters are different. In VideoDiT, $w_1, w_2, w_3=3$, $eta=0.5$. In AudioDiT, $w_1=2, w_2=10, w_3=2$, $eta=0.2$.

\begin{figure}[!t]
    \centering
    \includegraphics[width=1.0\textwidth]{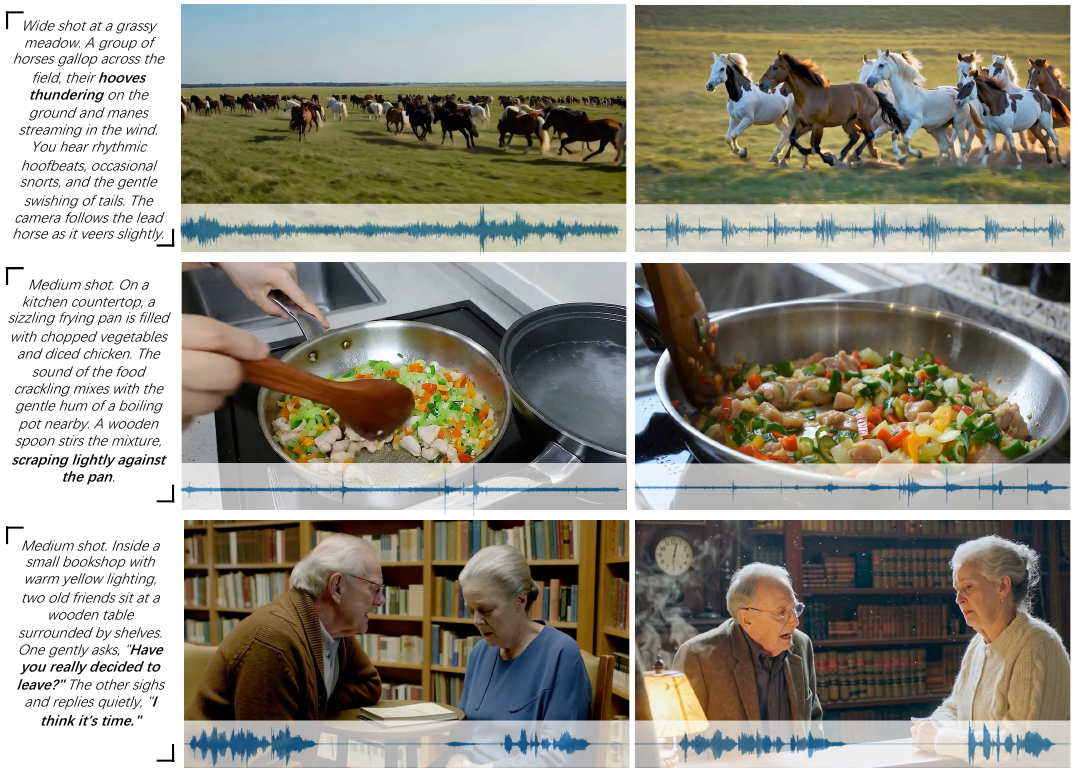}
    \caption{
        Qualitative comparison of generation results with and without our Prompt Engineering (PE) strategy. For each example, the \textbf{left} shows the output with structured prompt transferred from the raw user prompt, while the \textbf{right} is the result after PE enhancement.
    }
    \label{fig:abl_pe}
\end{figure}

\textbf{Prompt Engineering.} The quality of generative models is heavily contingent on the alignment between the user's input distribution and the model's training data distribution. Raw user prompts are often sparse and ambiguous, leading to suboptimal generation results. Instead of training a separate re-writer, we leverage the emergent capabilities of off-the-shelf LLMs~\citep{comanici2025gemini} to perform zero-shot prompt enhancement. Our strategy focuses on bridging the domain gap by mapping user inputs to the structured schema used during training. Specifically, we instruct the LLM to explicitly hallucinate missing details across multiple axes: for vision, it enriches descriptions of camera perspective, lighting, and background textures; for audio, it elaborates on environmental ambience and specific sound textures. This ensures the prompt lies within the model's high-density probability region, yielding more consistent performance.

To further bias the generation towards high aesthetic fidelity, we employ a quality steering mechanism. During the prompt rewriting phase, we provide the LLM with few-shot examples of captions corresponding to varying quality tiers (e.g., ``High Quality", ``Medium Quality"). By explicitly conditioning the rewriting process on a ``High Quality" token, we guide the LLM to adopt the linguistic style and vocabulary associated with our premium training samples. This effectively steers the downstream DiT to sample from the high-fidelity sub-manifold of the latent space.

A unique challenge in audio generation is the ambiguity of sound descriptions (e.g., ``loud noise" vs. ``explosion"). To ensure acoustic clarity, we propose a Retrieval-Augmented Refinement pipeline. We first construct a dense vector index of all unique sound effect descriptions present in our training corpus using FAISS~\citep{johnson2019billion}. During inference, we extract sound events from the user's prompt and query this database for the nearest neighbors based on cosine similarity. If a retrieval match exceeds a similarity threshold of $\tau > 0.85$, we replace the user's generic term with the retrieved canonical description. This substitution aligns the inference prompt with the specific acoustic concepts the model has mastered, significantly enhancing the distinctiveness and realism of the generated sound effects. 

As illustrated in Figure~\ref{fig:abl_pe}, the efficacy of our prompt engineering strategy is empirically evident across both modalities. Visually, the generated content exhibits a substantial elevation in aesthetic quality, characterized by sharper subject prominence and more coherent composition. This confirms that our approach effectively activates the model's latent capacity for high-aesthetic generation, successfully steering the output towards the distribution of premium data. Acoustically, the sound events are rendered with significantly greater definition and presence. For instance, the rhythmic cadence of horse hooves becomes strictly synchronized and distinct, while fine-grained sounds, such as the crisp impact of a wooden spoon against a pan, are synthesized with high fidelity. These qualitative improvements validate that aligning user prompts with the model's training schema is crucial for unlocking its full generative potential.

%% file: content/5_role_playing_animate.tex
\section{Role-Playing Animate}
\label{sec:roleplaying}

The recent emergence of advanced video generation models (e.g., Sora2) has demonstrated the feasibility of high-fidelity character animation, enabling the animation of single or multiple human subjects by referencing character images, and placing them in arbitrary scenarios while maintaining strict identity consistency. By integrating reference-guided learning with supervised fine-tuning (SFT) in a joint training paradigm, the enhanced \method not only ensures rigorous visual identity preservation of characters but also achieves precise alignment between character movements and synthesized audio streams.

\subsection{Multi-Reference Conditioning Training}
\method integrates reference images directly into the latent space of the DiT (Diffusion Transformer) backbone. Reference images (single or multiple) are first encoded through the VAE encoder. The resulting latent representations are then prepended to the input latent sequence of the target video. This allows the model to attend to the reference features throughout the denoising process via the self-attention mechanism.

A significant challenge in prepending reference frames is that the model may interpret them as the starting frames of a continuous video sequence, leading to continuous transition effects between the reference frames and the generated video. To resolve this, we introduce Ref-Temporal PE Offset strategy. We apply a negative shift to the temporal indices of the reference frames. This ensures that the model perceives the reference images as ``contextual anchors" rather than part of the temporal flow of the generated video. The temporal index $T_{ref}$ for the $k$-th reference image is calculated as:

\begin{equation}
T_{ref}(k) = -\phi \cdot k, \quad k \in \{1, \dots, K\}
\end{equation}

Where $K$ is the number of reference images, $\phi$ is the offset base coefficient. By shifting the reference tokens to a negative temporal space (e.g., -10, -20), we exploit the Ref-Temporal PE Offset to create a distinct separation between ``identity guidance" and ``temporal progression."

\begin{figure}[t]
    \centering
    \includegraphics[width=1.0\textwidth]{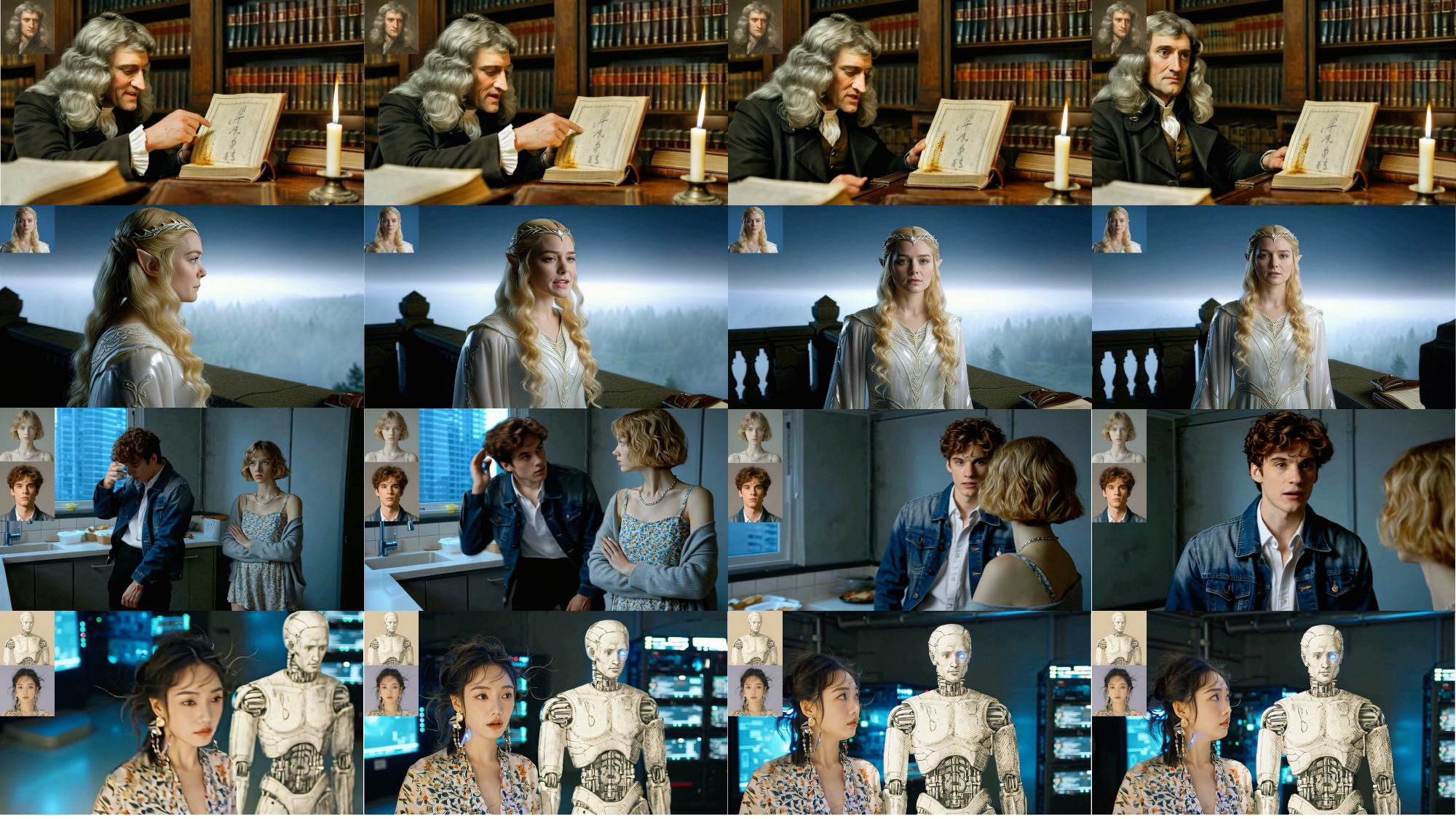}
    \caption{
        Role-playing animate examples using \method, with reference images and corresponding generated video frames (text prompts omitted). The left column shows the reference images, and the right four columns are sampled frames from generated subject-consistent videos.
    }
    \label{fig:ref_sampled_frames}
\end{figure}

\subsection{Multi-Reference Inference}
The inference process for Role-Playing Animate differs significantly from standard Text-to-Video-Audio (T2VA) tasks. While standard inference relies on a single Classifier-Free Guidance (CFG) scale to mediate between the text prompt and the unconditional output, the multi-reference task introduces a second conditioning dimension: the Visual Identity. To achieve high-fidelity character animation that respects both the narrative (text) and the character (reference images), we implement a Dual-Conditioning CFG strategy. We decouple the guidance into two distinct scales: Text CFG ($s_{txt}$) and Reference Image CFG ($s_{ref}$). The final noise prediction is formulated as follows:

\begin{equation}
\begin{aligned}
\hat{\epsilon}_{\theta} &= \epsilon_\theta(z_t, \emptyset, \emptyset) + s_{txt} \cdot \big(\epsilon_\theta(z_t, c_{txt}, \emptyset) - \epsilon_\theta(z_t, \emptyset, \emptyset)\big) \\
& \quad + s_{ref} \cdot \big(\epsilon_\theta(z_t, c_{txt}, c_{ref}) - \epsilon_\theta(z_t, c_{txt}, \emptyset)\big)
\end{aligned}
\label{eq:epsilon_hat}
\end{equation}

Where:$\epsilon_\theta(z_t, \emptyset, \emptyset)$ is the fully unconditional prediction (null text, null reference).$\epsilon_\theta(z_t, c_{txt}, \emptyset)$ is the text-only prediction, which guides the general motion and scene layout.$\epsilon_\theta(z_t, c_{txt}, c_{ref})$ is the jointly conditioned prediction, which injects the specific identity and texture of the reference images into the existing text-guided framework.

%% file: content/6_bench.tex
\section{Benchmark Evaluation}

\subsection{Comprehensive Evaluation}

\begin{figure}[!t]
    \centering
    \includegraphics[width=\textwidth, trim=0 60 0 105, clip]{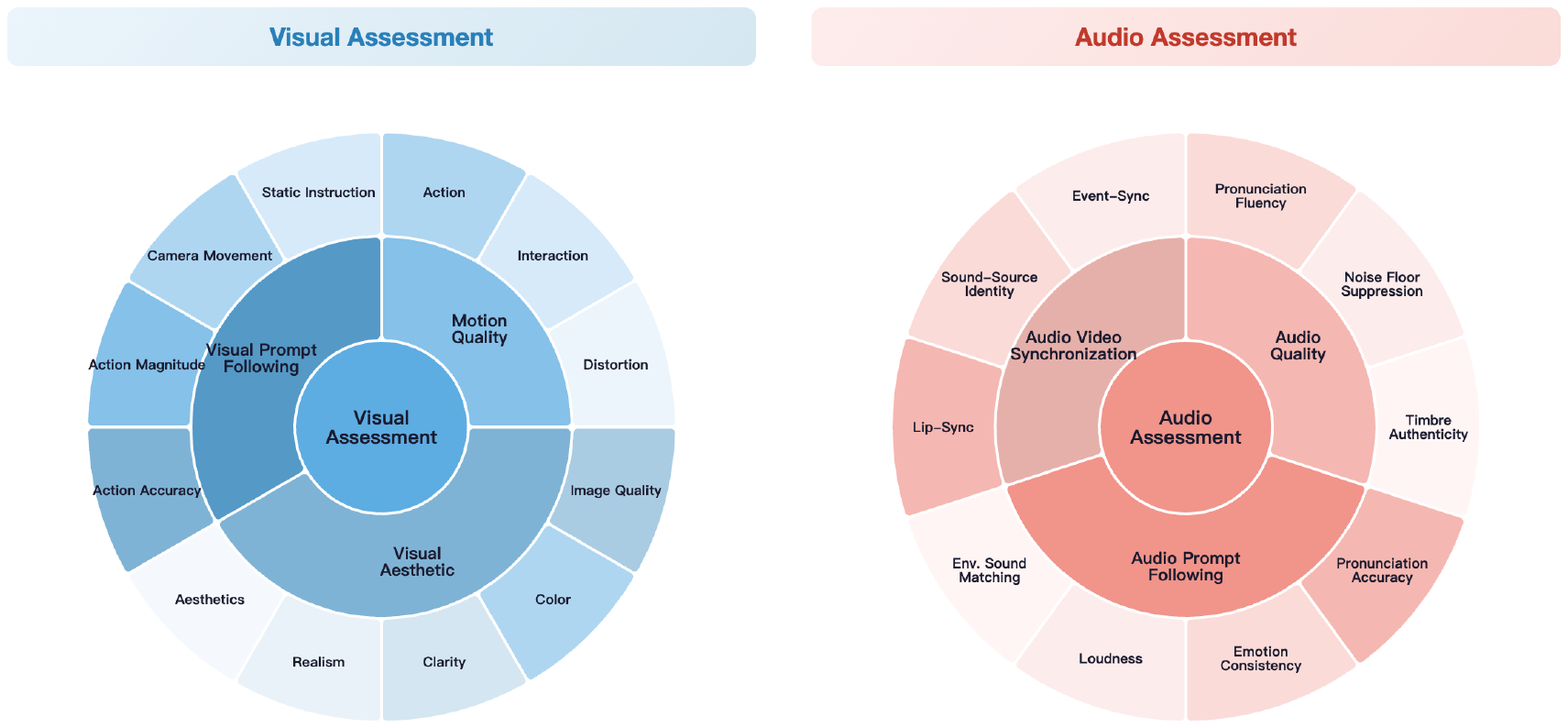}
    \caption{
        Detail evaluation metrics of Alive-Bench 1.0. \method. 
    }
    \label{fig:bench_metrics_detail}
\end{figure}

Our benchmark is designed to offer clear advantages in representativeness, robustness, and interpretability by: (i) covering a broad and compositional prompt space to characterize real-world capability boundaries, (ii) adopting prompts that closely resemble user inputs to reduce distribution shift between evaluation and deployment, and (iii) employing fine-grained, multi-dimensional metrics to enable diagnostic assessment of visual, audio, and audio video synchronization.
\paragraph{Prompt Diversity.}
We construct prompts with extensive diversity across: multiple subjects (e.g., humans, animals, means of transportation, musical instruments, everyday objects, etc.), multiple languages (Chinese and English), multiple interaction paradigms (single-speaker vs. multi-party dialogue), multiple scenes (e.g., urban, natural, home, sports venues, outer space, etc.), and diverse cinematography and styles (varied shooting techniques, stylistic renderings, surreal settings). These design choices bring the benchmark closer to the long-tail distribution of real-world content production.
Prior evaluations often over-emphasize the primary subject, while key factors that determine whether outputs feel real frequently arise from background context, soundscape, physical interactions, and global coherence. By incorporating scene/style/camera variations, we impose constraints along multiple axes, requiring a model to satisfy 
subject plausibility, scene plausibility, interaction plausibility and audio-visual synchronization together.
This mitigates the common failure mode where the subject appears correct but the overall result is unnatural.
\paragraph{User-Like Inputs.}
Real user requests are typically multi-condition compositions, e.g.,
multiple subjects, multi-turn dialogue, specified camera language, specified scene and specified acoustic attributes together. Accordingly, our prompt design systematically probes whether models exhibit characteristic degradations under compositional constraints, including multi-party dialogue: speaker ID drift and turn misalignment, non-human subjects: physical/acoustic inconsistencies (e.g., mismatched vehicle timbre; instrument-playing actions misaligned with produced sounds), scene-driven audio: missing or incorrect ambience (e.g., indoor reverberation, outdoor wind noise, spatial cues), camera/style shifts: fluctuations in visual-motion quality and degraded aesthetic coherence.
This setup evaluates not only isolated capabilities, but also stability under complex, realistic conditions.
\paragraph{Fine-Grained and Diagnostic Metrics.} Our evaluation spans six major categories—motion quality, visual aesthetic, visual prompt following, audio quality, audio prompt following and audio video synchronization—and further decomposes them into 22 fine-grained dimensions. The key benefit is not merely broader coverage, but the provision of interpretable, localizable, and optimizable signals.
Models frequently exhibit trade-offs such as natural audio but poor A/V sync or strong motion but weak aesthetics. Multi-dimensional evaluation makes such trade-offs explicit rather than obscured by a single aggregate score. Moreover, aggregate metrics provide limited guidance for iteration, whereas fine-grained dimensions enable attribution of errors to specific components.
These metrics naturally support: regression tracking across model versions, ablation studies over data and alignment strategies and capability profiling across scenarios and conditions.
\paragraph{Evaluation Metrics.}
We define a comprehensive set of metrics for our human evaluation, organized into six main categories, motion quality, visual aesthetic, visual prompt following, audio quality, audio prompt following and audio video synchronization. Figure~\ref{fig:bench_metrics_detail} illustrates the detailed categorization of metrics.

\begin{itemize}
    \item \textbf{Motion Quality.} This category assesses the physical realism and coherence of movement. This includes: \textit{Action}, which evaluates if the motion is natural and smooth, conforming to the inherent laws of movement; \textit{Interaction}, which checks if interactions (between subjects, subject-object, or between objects) adhere to physical laws; and \textit{Distortion}, which determines if the subject exhibits artifacts like noise, frame loss, or blurring, and ensures the subject remains consistent throughout.
    \item \textbf{Visual Aesthetic.} This category evaluates the aesthetic and technical quality of the image itself. Key aspects are: \textit{Image Quality}, covering texture, lighting, and the presence of visual artifacts such as flickering or overexposure; Color, for the appropriateness and appeal of the color scheme; \textit{Clarity}, measuring image sharpness; \textit{Realism}, determining if the visual output is sufficiently realistic (or stylistically consistent for non-realistic styles); and \textit{Aesthetics}, for the overall visual appeal. 
    \item \textbf{Visual Prompt Following.} This category measures how faithfully the generated video adheres to the user's text prompt. This is broken down into: \textit{Action Accuracy}, checking the correctness of actions, temporal sequence, and direction; \textit{Action Magnitude}, assessing the amplitude and speed of actions; \textit{Camera Movement}, evaluating camera operations like panning, zooming, and tilting; and \textit{Static Instruction}, judging subjects' appearances, numbers, spatial relationships, etc., and background are correct.
    \item \textbf{Audio Quality.} This category assesses the clarity and naturalness of synthesized audio. This includes: \textit{Pronunciation Fluency}, which evaluates the naturalness of speech in terms of rhythm and coherence, focusing on whether speaking rate, pauses, and speech flow conform to natural pronunciation rhythm. \textit{Noise Floor Suppression}, which assesses the degree of control over residual noise levels (e.g., electrical hum). \textit{Timbre Authenticity}, which evaluates whether the timbre is natural and full-bodied, and whether it matches the subject specification and the intended scenario.
    \item \textbf{Audio Prompt Following.} This category evaluates how accurately the generated audio complies with the specified vocal and acoustic constraints. Key aspects are: \textit{Pronunciation Accuracy}, degree to which syllables and words are produced with correct phonemes, clear articulation, and high intelligibility, without phoneme substitutions, muffled/unclear sounds, slurring, or omissions. \textit{Emotion Consistency}, degree to which expressed emotional type and intensity match the semantic and contextual cues. \textit{Loudness}, degree to which the loudness conforms to the prompt and the target context. \textit{Env. Sound Matching}, which assess the alignment between the background ambience and the scene specified in the instruction.
    \item \textbf{Audio Video Synchronization.} This category measures the coherence between audio and video. This is broken down into: \textit{Lip-Sync}, degree of temporal consistency between speech and lip/facial movements. \textit{Sound-Source Identity}, which evaluates whether the identity of the sound source is consistent with the prompt. \textit{Event-Sync}, which measures non-speech sound events (e.g., animal vocalizations, instrument sounds, vehicle noises) are temporally aligned with the video.
\end{itemize}

\begin{figure}[!t]
    \centering
    \includegraphics[width=\textwidth, trim=0 60 0 40, clip]{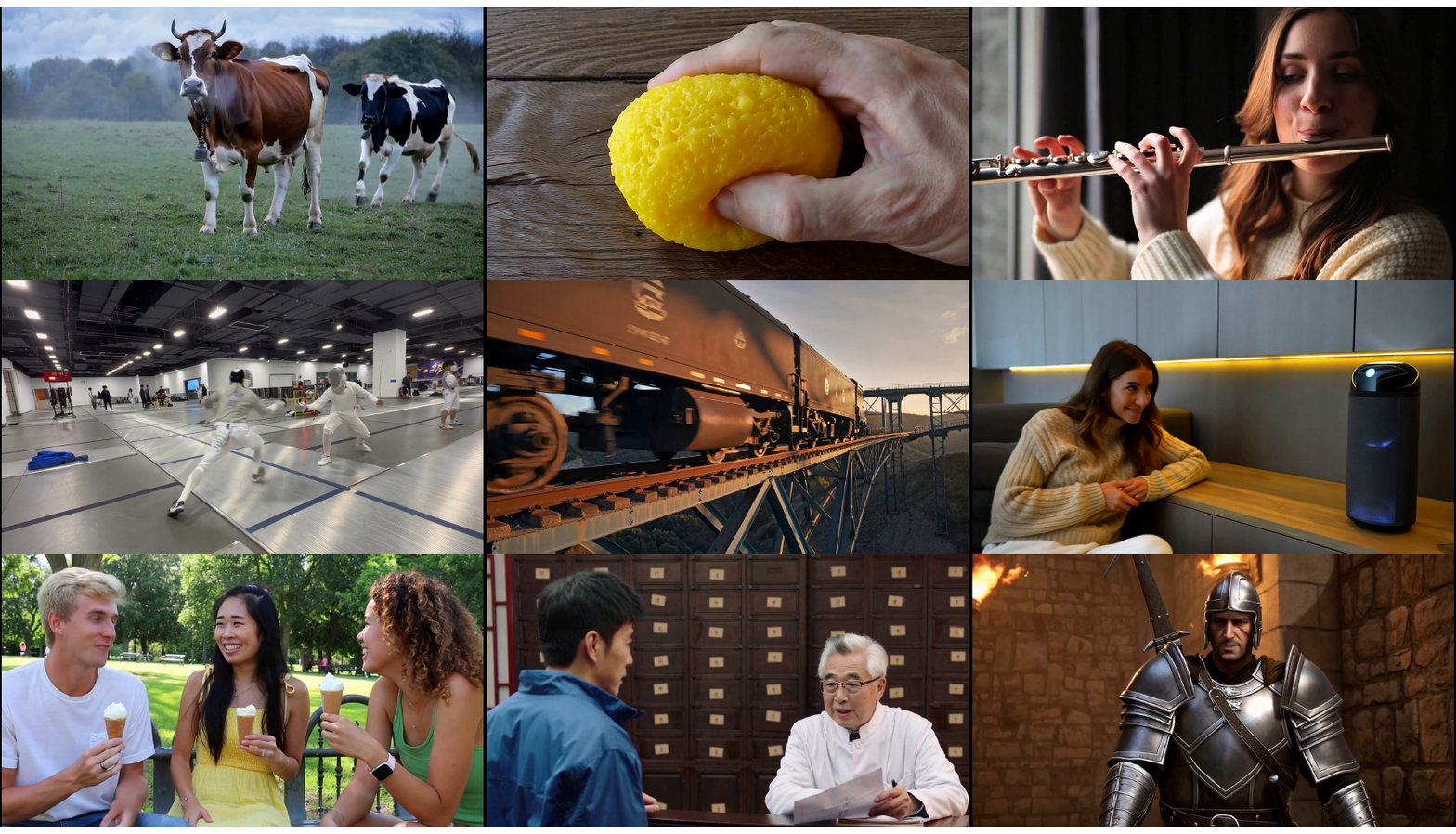}
    \caption{
        Some video examples of the Alive-Bench 1.0 generated by \method. 
    }
    \label{fig:common_show}
\end{figure}

\paragraph{Evaluation Benchmarks.}
To apply these metrics comprehensively we introduce \textbf{Alive-Bench 1.0}, which is a broad, general-purpose benchmark consisting of 264 samples that covers a wide range of scenarios, including single-person speech, multi-people conversations, sports, daily activities, animals, means of transportation, surreal scenes, etc. Figure~\ref{fig:common_show} presents several generated examples for the benchmark.

\begin{figure}[!t]
    \centering
    \includegraphics[width=\textwidth, trim=0 60 0 60, clip]{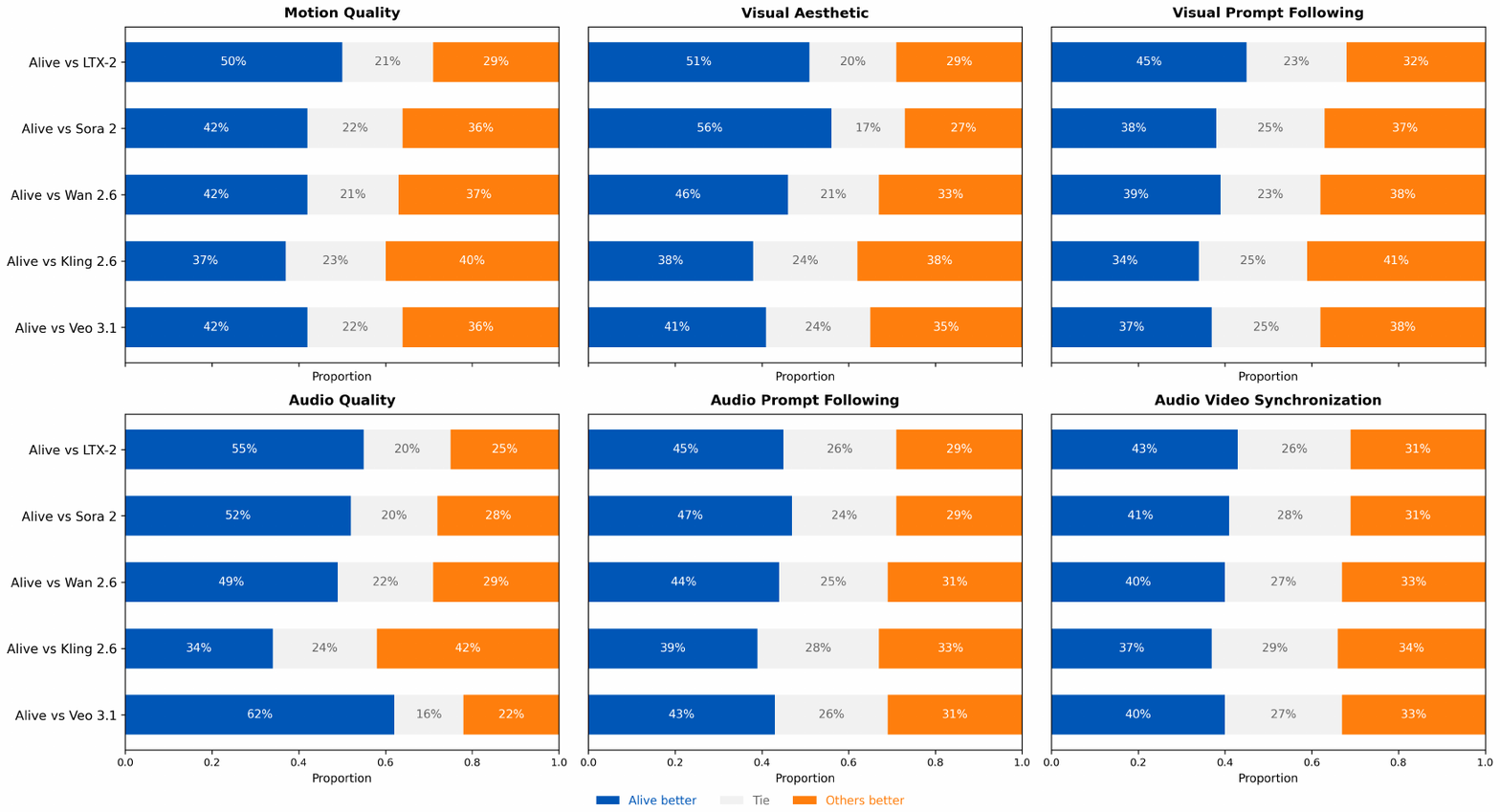}
    \caption{
        \textbf{Alive-Bench 1.0 human evaluation results comparing our model (Alive) with leading competitors.}
        Users were presented with side-by-side video comparisons and asked to choose which was better or if they were tied. The stacked bar charts show the proportion of user preference votes across different dimensions.
    }
    \label{fig:human_evaluation_combined}
\end{figure}

\paragraph{Human Evaluation.}
We conducted extensive two-round human evaluations to benchmark our model’s performance against leading competitors (Veo 3.1, Kling 2.6, Wan 2.6, Sora 2, and LTX-2). Using a side-by-side comparison protocol, human raters were presented with pairs of videos generated by our model (Alive) and a competing model in a blinded manner (i.e., raters were not informed which video came from which model) and were asked to indicate which video was superior. Raters’ preferences were determined according to the comprehensive evaluation criteria defined above. Figure~\ref{fig:human_evaluation_combined} shows the results.

\begin{itemize}
    \item \textbf{Motion Quality.} Alive's motion quality is better than Veo 3.1, Wan 2.6, Sora 2 and LTX-2, and is slightly behind Kling 2.6.
    \item \textbf{Visual Aesthetic.} Alive shows strong visual aesthetic ability. Alive significantly outperforms Veo 3.1, Wan 2.6, Sora 2 and LTX-2 on visual aesthetic, and performs on par with Kling 2.6.
    \item \textbf{Visual Prompt Following.} Alive demonstrates a clear advantage over LTX-2 on visual prompt following, and achieves near-parity with Veo 3.1, Wan 2.6 and Sora 2. And Alive underperforms Kling 2.6.
    \item \textbf{Audio Quality.} Alive exceeds Veo 3.1, Wan2.6, Sora 2 and LTX-2 by a wide margin, while falls short of Kling 2.6 on audio quality.
    \item \textbf{Audio Prompt Following.} Alive outperforms all five competitors by a substantial margin on audio prompt following.
    \item \textbf{Audio Video Synchronization.} Alive leads the five competing models with the strongest audio video synchronization ability.
\end{itemize}
Across all metrics, Alive ranks at or near the top, indicating a well-balanced capability profile rather than a single-metric advantage. Alive is the best on audio prompt following and audio video synchronization, outperforming other competitors by a notable margin. This indicates a strong advantage in cross-modal understanding and alignment, particularly in faithfully reflecting audio instructions and maintaining tight timing correspondence between audio events and visual content. Overall, the evaluation results imply that Alive is especially well-suited for sound-critical video generation scenarios, where accurate audio conditioning and synchronization are primary requirements, while still delivering near-best-in-class motion, visual instruction adherence, and audio fidelity.

\subsection{Role-Playing Animate Evaluation}

\paragraph{Evaluation Metrics.}
Besides the six main categories defined above, we additionally introduce a category dedicated to Role-Playing Animate.

\begin{itemize}
    \item \textbf{Reference-Character Faithfulness.} This category assesses the generation quality of the reference character. This includes: \textit{Character Preservation}, which measures the extent to which the generated content preserves the reference person's visual identity. \textit{Fusion Naturalness}, which evaluates how naturally the reference character blends with the surrounding environment in the generated video.
\end{itemize}

\begin{figure}[!t]
    \centering
    \includegraphics[width=\textwidth, trim=20 60 0 40, clip]{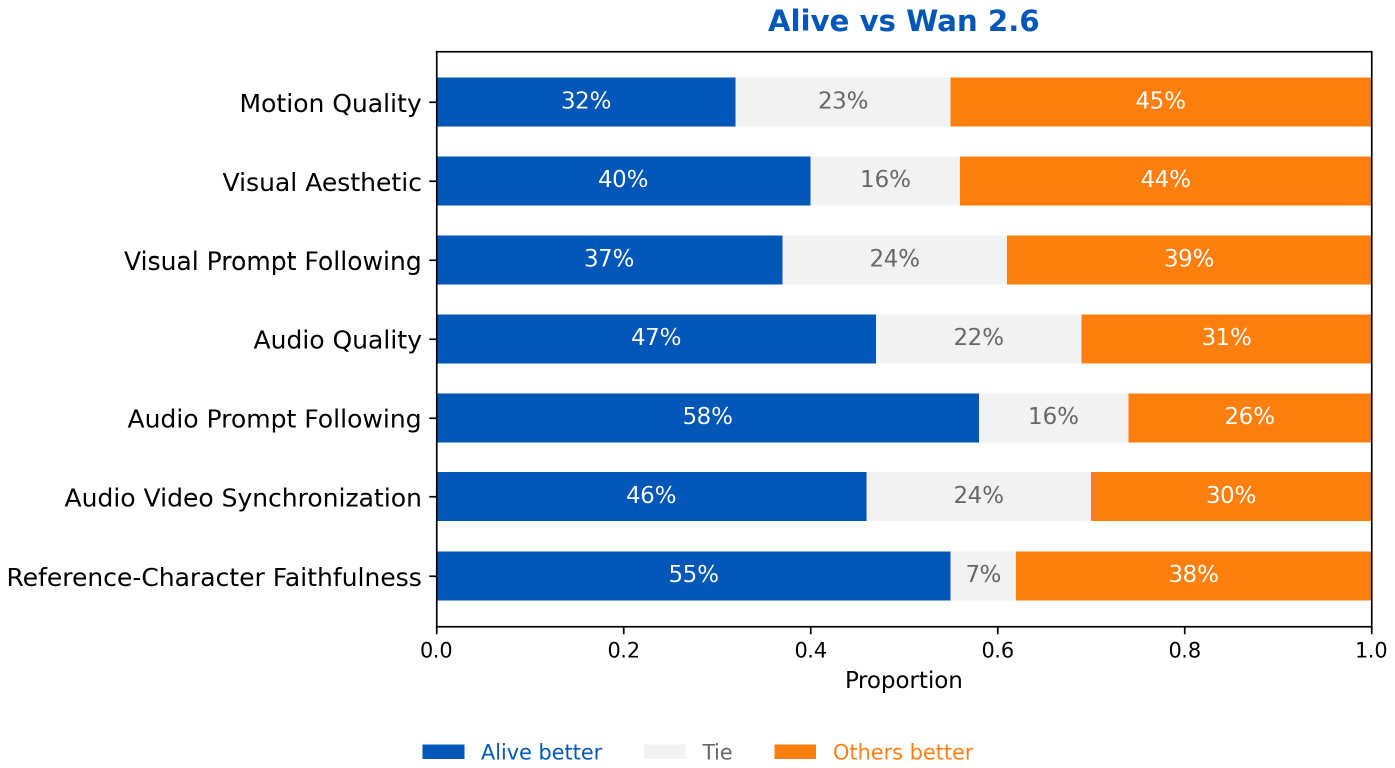}
    \caption{
        \textbf{Alive-Bench 1.0 reference-character sub-benchmark human evaluation results comparing our model (Alive) with Wan 2.6.}
        Users were presented with side-by-side video comparisons and asked to choose which was better or if they were tied. The stacked bar charts show the proportion of user preference votes across different dimensions.
    }
    \label{fig:human_evaluation_role_playing}
\end{figure}

\paragraph{Evaluation Benchmarks.} We construct the Alive-Bench 1.0 reference-character sub-benchmark, comprising 90 prompts: 29 with single-person reference images and 61 with two-person reference images. Figure~\ref{fig:human_evaluation_role_playing} shows the results.

Alive’s reference-character faithfulness is substantially higher than Wan 2.6, indicating that our model excels at the most critical capability for Role-Playing Animate task—faithfully generating the reference character. In addition, Alive consistently outperforms Wan 2.6 on Audio Quality, Audio Prompt Following and Audio Video Synchronization, suggesting stronger audio expressiveness and audio-conditioned generation. Our model performs on par with Wan 2.6 in visual prompt following, while trailing slightly in motion quality and visual aesthetic.

%% file: content/8_conclusion.tex
\section{Conclusion and Limitation}
\label{sec:conclusion}
\paragraph{Conclusion.}
We introduce \method, a rectified flow Transformer framework that unifies audio and video generation tasks and achieves high-performance generation results. Through a robust audio-video data curation pipeline and detailed benchmark evaluation metrics, \method demonstrates superior performance across benchmarks, especially in audio quality and audio-video synchronization. We hope that the insights and practical recipes provided in this work will support the community in further advancing the state of audio-video generation technology. 

\paragraph{Limitation.}

There are certain limitations. After applying the audio-video data curation pipeline to a fixed video pool, the available data volume is smaller than that of video generation model. Consequently, the model may fail to generate accurate visual representations for concepts with limited representation in the training data.

%% file: content/9_authors.tex
\section{Contributions and Acknowledgments}
\label{contributions}
\textbf{Core Contributors: }

Ying Guo,  Qijun Gan, Yifu Zhang, Jinlai Liu, Yifei Hu, Pan Xie, Bingyue Peng, Zehuan Yuan

\textbf{Contributors: }

Dongjun Qian, Yu Zhang, Ruiqi Li, Yuqi Zhang, Ruibiao Lu, Xiaofeng Mei, Bo Han, Xiang Yin 

\textbf{Acknowledgments: }

Changhao Pan, Tianyun Zhong, Qinxin Wu,  Ruoyu Guo, Ge Bai, Xin Ji,  Chongxi Wang,  Hao Yang, Chuang Lin, Hui Wu,  Bobo Zeng, Yina Tang, Dongyang Wang, Dongdong Yang, Xiaoxiao Qin, Colin Young, Fangzhou Ai, Mingyang Zou, Ke Lei

\clearpage